\documentclass{article}

\usepackage[T1]{fontenc}
\usepackage{times}
\usepackage{latexsym}
\usepackage[utf8]{inputenc}
\usepackage{pifont}
\usepackage{amssymb}
\usepackage{microtype}
\usepackage{wrapfig}
\usepackage{graphicx}
\usepackage{subfigure}
\usepackage{multicol}
\usepackage{multirow}
\usepackage[most]{tcolorbox}
\usepackage{soul}
\usepackage{subcaption}
\usepackage{inconsolata}
\usepackage{listings}
\definecolor{codegray}{rgb}{0.9,0.9,0.9}
\newtcolorbox[list inside=prompt,auto counter,number within=section]{prompt}[1][]{
    colbacktitle=black!60,
    fonttitle=\small,
    coltitle=white,
    fontupper=\footnotesize,
    boxsep=4pt,
    left=0pt,
    right=0pt,
    top=0pt,
    bottom=0pt,
    boxrule=1pt,
    #1,
}
\usepackage{booktabs} 

\usepackage{hyperref}

\newcommand{\name}[0]{OWLViz}
\newcommand{\datasetsize}[0]{248}

\usepackage[accepted]{icml2025}

\usepackage{amsmath}
\usepackage{amssymb}
\usepackage{mathtools}
\usepackage{amsthm}

\usepackage[capitalize,noabbrev]{cleveref}

\theoremstyle{plain}

\theoremstyle{definition}

\theoremstyle{remark}

\usepackage[textsize=tiny]{todonotes}

\icmltitlerunning{ICML 2025 Workshop on
Multi-Agent Systems in the Era of Foundation Models: Opportunities, Challenges and Futures}

\begin{document}

\twocolumn[
\icmltitle{OWLViz: An Open-World Benchmark for Visual Question Answering}



\icmlsetsymbol{equal}{*}

\begin{icmlauthorlist}
\icmlauthor{Thuy Nguyen}{reasoning}
\icmlauthor{Dang Nguyen}{maryland}
\icmlauthor{Hoang Nguyen}{ptit}
\icmlauthor{Thuan Luong}{ptit}
\icmlauthor{Franck Dernoncourt}{adobe}
\icmlauthor{Long Dang Hoang}{ptit}
\icmlauthor{Viet Dac Lai}{adobe}
\end{icmlauthorlist}

\icmlaffiliation{reasoning}{Reasoning Foundation, USA}
\icmlaffiliation{ptit}{Posts and Telecommunications Institute of Technology, Viet Nam}
\icmlaffiliation{adobe}{Adobe Research, USA}
\icmlaffiliation{maryland}{University of Maryland, USA}

\icmlcorrespondingauthor{Long Dang Hoang}{longdh@ptit.edu.vn}

\icmlkeywords{Machine Learning, ICML}

\vskip 0.3in
]



\printAffiliationsAndNotice{}  

\begin{abstract}
We present a challenging benchmark for the Open WorLd VISual (OWLViz) question answering benchmark. \name~ presents short queries that require integrating multiple capabilities, including common-sense knowledge, visual understanding, web exploration, and specialized tool usage. While humans achieve 69.2\% accuracy on these intuitive tasks, even state-of-the-art VLMs struggle, with the best model, Gemini, achieving only 27.09\% accuracy. Current tool-calling agents and GUI agents, which rely on limited vision and vision-language models as tools, perform even worse. This performance gap reveals significant limitations in multimodal systems' ability to select appropriate tools and execute complex reasoning sequences, establishing new directions for advancing practical AI research.
    
\end{abstract}

\section{Introduction}

Large Vision Language Models (VLMs) have recently demonstrated impressive visual understanding and reasoning capabilities across numerous tasks \cite{liu2024visual}. 
Equipped with these advanced capabilities, VLMs are rapidly surpassing existing AI benchmarks \cite{chen2024expanding}. As the AI community pursues increasingly challenging problems, benchmark tasks are shifting from conventional fundamentals toward more human-centric challenges \cite{hu-etal-2024-viva, roger2023towards, brohan2023rt}. These human-centric tasks typically demand broader world knowledge, situation awareness, and more complex reasoning. 


While there are many benchmarks for VQA that have been created, existing efforts face several notable problems. Many early benchmarks only focus on visual understanding tasks such as entity detection and entity attribution \cite{ren2015cocoqa,malinowski2014daquarqa,balanced_vqa_v2}. As a result, these benchmarks can only be used to evaluate the VLM's grounding capability, while many questions require additional capabilities such as reasoning. Recent benchmarks have shifted the focus towards complex questions where compositional reasoning  and spatial reasoning are evaluated \cite{hudson2019gqa, gao2023cric, bitton2023visit}. 
 Even though tool calling capability was investigated in a close-environment \cite{liu2023agentbench,li-etal-2023-api,patil2024gorilla}, these datasets only addressed a limited set of tools. As a result, these benchmarks fail to fully capture the complexity of real human queries that often require interpreting complex visual contexts, combining heterogeneous sources of knowledge,  and reasoning over long, multi-step chains of actions \cite{mialon2023gaia}.
As such, current evaluations do not sufficiently evaluate a model’s ability to act as an agent that can meaningfully invoke and coordinate tools to solve practical, open-ended questions grounded in the visual world. 

\begin{figure*}[!h]
    \centering
    \subfigure[
        {\bf Question:} {\it ``How many people are visible on the left side of the white line that cuts across the photo? Provide a numeric answer.''} \newline 
        {\bf Answer:} 2  \newline 
        {\bf Skills:} using external API, human recognition. \newline 
        {\bf Difficulty level:} 1.
    ]{
        \includegraphics[height=3.7cm]{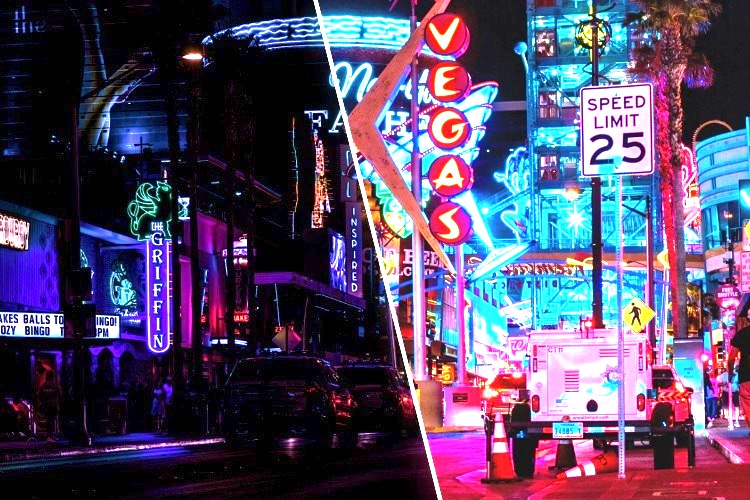}
    }
    \hfill
    \subfigure[
        {\bf Question:} {\it ``How many umbrellas have 3 or more colors? Provide a numeric answer.''} \newline 
        {\bf Answer:} 2. \newline
        {\bf Skills:} object recognition, attribute identification, counting, object detection. \newline 
        {\bf Difficulty level:} 2.
    ]{
        \includegraphics[height=3.7cm]{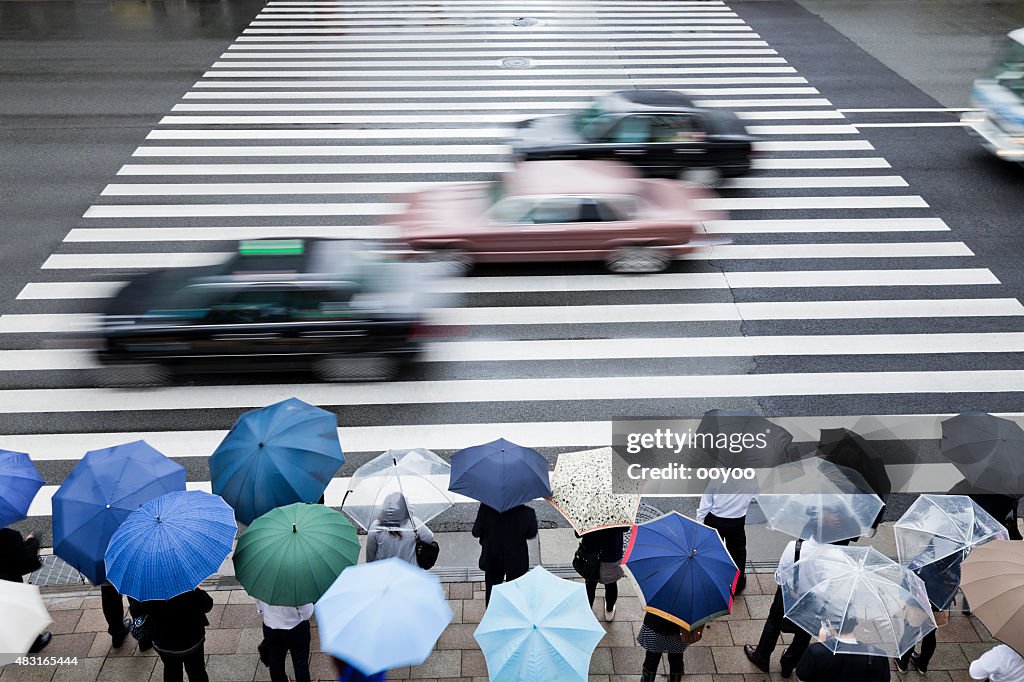}
    }
    \hfill
    \subfigure[
        {\bf Quesstion:} {\it ``This is in Fairfax, Virginia. What is the name of the road shown in the photo?''} \newline 
        {\bf Answer}:  Shadowridge Dr; Shadowridge drive; Shadowridge.
        \newline 
        {\bf Skills:} OCR, knowledge search, knowledge retrieval, GUI, comparison, spatial relationships. \newline 
        {\bf Difficulty level:} 3.
    ]{
        \includegraphics[height=3.7cm]{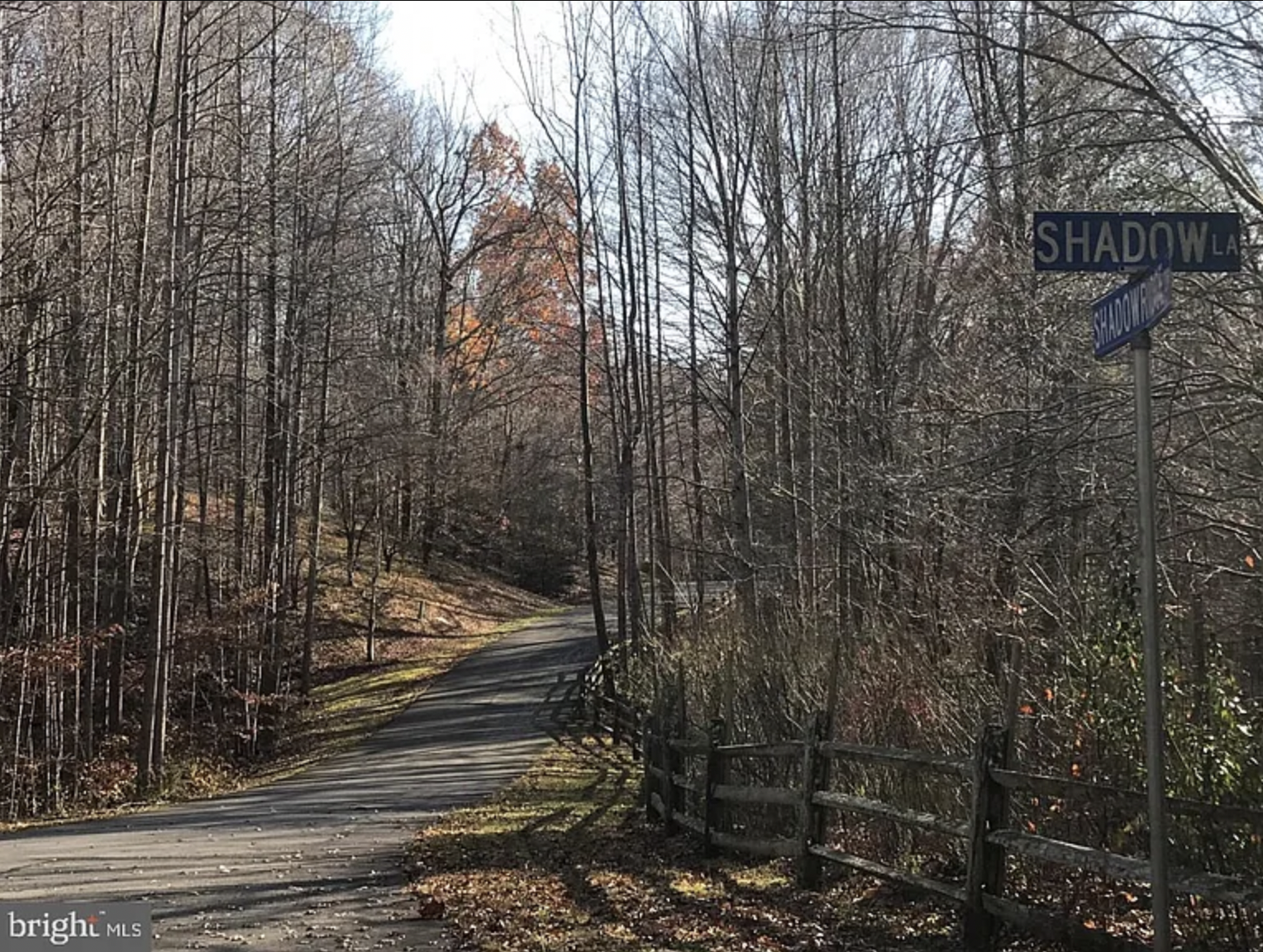}
    }
    \caption{ {\bf Examples of the three core challenges in our OWLViz  dataset}. (a) Challenging visual conditions require image enhancement or specialized recognition tools to count people on a white line in a low-contrast night scene. (b) Complex reasoning tasks demanding object detection, attribute identification, and precise counting of multi-colored umbrellas in a dynamic street scene. (c) Knowledge-intensive queries require internet exploration and external data retrieval to identify specific locations based on minimal visual cues.}
    \label{fig:examples}
\end{figure*}

To address this gap, we introduce \name, a novel benchmark dataset specifically designed to evaluate vision-language models' ability to utilize tools in complex, multimodal reasoning tasks. \name~ necessitates three distinct skill sets that challenge both VLMs and agentic systems. \textbf{First}, the dataset incorporates visually degraded inputs featuring low brightness, poor contrast, or blur, scenarios where visual enhancement tools may be required to improve image quality for accurate processing. \textbf{Second}, tasks demand sophisticated reasoning capabilities to solve complex problems involving counting, projection, and measurement operations. \textbf{Third}, certain challenges require models to explore the internet and retrieve external data to answer questions based on minimal visual cues. Figure \ref{fig:examples} illustrates representative examples of these challenging scenarios.

\name~ introduces a benchmark for Agentic AI Assistant featuring \datasetsize~ carefully annotated questions and answers where big proprietary VLM models failed to answer. 
\name~ dataset is easy to understand, challenging for both human and AI  and featuring extensive tool-use skills. Yet, evaluation on this dataset is simple and can be done automatically.

Despite their success in many visual grounding tasks, even the best-performing VLMs demonstrated surprisingly poor performance on our benchmark. Most models achieved less than 20\% accuracy in exact-match evaluation and below 30\% in LLM-match evaluation, while college students easily completed these tasks with 69.2\% accuracy in under one minute. 
\name~presents novel challenges and opportunities for advancing research in open-world visual understanding. We keep our dataset private to minimize data contamination. Additional information on how to access the data is available upon request.

\section{Related work}
\paragraph{VQA Dataset.}

Early visual question answering datasets primarily focused on evaluating fundamental visual abilities such as visual entity recognition and entity attribution identification  \cite{ren2015cocoqa,malinowski2014daquarqa}, establishing foundational benchmarks for basic visual comprehension. As the field matured, later works shifted focus toward more sophisticated challenges, including object relational understanding  \cite{nntol2015vqa}, compositional reasoning involving complex spatial relationships  \cite{johnson2017clevr,hudson2019gqa}. These datasets introduced structured reasoning tasks that required models to understand not merely individual objects, but their interactions and hierarchical relationships within visual scenes. Recent studies have further expanded the scope by incorporating common-sense knowledge requirements  \cite{marino2019okvqa}. pushing models to integrate world knowledge with visual understanding. Despite these advances, current datasets remain quite limited in their coverage of real user queries, which often involve multi-step reasoning, require external tool usage, or demand interaction with challenging visual conditions that reflect authentic application scenarios. 

\paragraph{VLMs Dataset.}

The rapid development of state-of-the-art VLMs has led to significant advancements in vision understanding \cite{DBLP:conf/cvpr/HuSLVHLKF24, DBLP:journals/corr/abs-2405-20795, DBLP:conf/acl/ZhouLM024, le2024progressive}, task comprehension \cite{jing-etal-2024-faithscore}, reasoning \cite{Yue_2024_CVPR, dang2024sadl}, and tool-use capabilities \cite{liu2025llavaplus,nguyen2024dynasaur}. These capabilities pave the way for AI systems that can solve general human tasks    \cite{DBLP:journals/corr/abs-2403-09027}.
As a result, VLMs have surpassed AI benchmarks that were once considered highly challenging   \cite{mialon2023gaia, marino2019okvqa, hendrycks2020measuring}. In response, AI benchmarking efforts are shifting toward tasks that better reflect real-world applications   \cite{DBLP:conf/iclr/JimenezYWYPPN24, DBLP:journals/corr/abs-2411-04872} that incorporate human-centric considerations like ethical alignment and societal impact  \cite{hu-etal-2024-viva}.

\paragraph{Agentic Dataset.} Many agentic datasets for tool calling are dedicated to carefully curated environments with limited tool spaces   \cite{liu2023agentbench,li-etal-2023-api,patil2024gorilla}. where agents operate within predefined boundaries and constrained action sets. These controlled settings, while useful for systematic evaluation, fundamentally limit the assessment of agents' adaptability and generalization capabilities. Such datasets risk inadequately evaluating general-purpose agents that must navigate diverse, unpredictable real-world scenarios where tool selection and usage strategies cannot be predetermined.   Recent works have shifted towards open-domain, unlimited tools \cite{mialon2023gaia}. While GAIA is close to our work, its questions provide detailed plans to achieve the answer, hence only testing how well an agentic system is integrated with available tools rather than exploration and tool retrieval \cite{nguyen2024dynasaur}.

\section{Dataset Creation}

\name~is designed to evaluate models’ capabilities in comprehending image content and leveraging external tools to answer image-related questions. The dataset comprises \datasetsize~ human-designed and annotated questions, each associated with an image and an unambiguous ground truth answer, allowing exact-match evaluation.

This section presents our methodology for constructing a dataset that more effectively evaluates Visual Question Answering (VQA) systems' performance on open-world visual queries. The dataset development process encompasses five principal phases: (1) systematic image acquisition, (2) open-world question design, (3) establishment of reasoning paths to ensure reliable answer derivation, (4) comprehensive data annotation, and (5) implementation of standardized answer formatting protocols. 


\subsection{Image Acquisition}

Images were collected from a range of publicly accessible sources. For questions requiring only visual interpretation, we intentionally selected dense and detailed images to ensure sufficient visual complexity. For questions designed to prompt external searches, images were chosen to provide minimal but sufficient cues such as brand names, logos, or partial text while avoiding overexposure of information that could reduce the challenge of the task.

The five most frequent image sources are \textit{redfin.com}, \textit{istockphoto.com}, \textit{zillowstatic.com}, \textit{rdcpix.com}, \textit{shutterstock.com}, and \textit{pexels.com}. Together with 17 screenshots, these sources account for approximately half of the dataset. The 17 author-generated images were created in cases where direct image URLs were unavailable, such as screenshots taken from PDF reports, social media posts, or frames from YouTube videos.



\subsection{Questions Design} 

The dataset shares several design principles with GAIA, including (1) targeting questions that are conceptually simple yet practically useful and challenging for contemporary AI systems, (2) ensuring interpretability, (3) maintaining robustness against memorization, and (4) facilitating ease of evaluation.
However, \name~ distinguishes itself in several key aspects: (5) exclusive focus on images and (6) concise and practical. 

First, every question in \name~ is directly linked to a photograph. Questions can be answered using (i) image content alone, (ii) a combination of image content and metadata embedded within the image, or (iii) an integration of image content, metadata, and external knowledge inferred from visual cues. Unlike other datasets where questions can be excessively lengthy and unlikely to reflect the way people naturally inquire about information, our dataset prioritizes practicality. 

Second, \name~ is designed to capture questions that closely mirror the types of inquiries people naturally make when interpreting images in everyday contexts. These include tasks such as counting objects, identifying locations or addresses, checking property costs, extracting key details, or interpreting relationships between elements in a scene. This emphasis on practical, context-driven questions ensures the dataset's relevance to real-world applications and everyday problem-solving scenarios.

We develop an initial set of skills that was incorporated into the annotation web interface; the skill list includes three main categories: visual skills, reasoning skills, and tool skills (See Table \ref{tab:skill_taxonomy}). During the annotation, we allow annotators to add new skills if needed, which results in some less frequent skills as shown in Figure \ref{fig:skill_by_annotators}.

\begin{table*}[]
    \caption{Taxonomy of skills required for our dataset, categorized into visual, reasoning, and tool-based capabilities. Each skill is accompanied by a functional description indicating the specific capability it represents in the context of visual question answering tasks. }
    \resizebox{\linewidth}{!}{
    \begin{tabular}{c|l|l}
        \toprule
        & \multicolumn{1}{c|}{\bf Skill} & \multicolumn{1}{c}{\bf Description} \\
        \midrule
        \multirow{7}{*}{\rotatebox{90}{\bf Visual}}
        & human recognition & Identify humans and their locations in the image \\ 
        & object recognition & Identity non-human objects and their location in the image \\
        & identify 2 endpoints &  Identify two specific endpoints of an object in an image \\
        & object detection &  Identifies what objects are present and where they are located by using a bounding box  \\
        & object segmentation &  Provides pixel-level understanding of what and where all objects are  \\
        & spatial relationships &  Understands 3D spatial relations between objects in the image.\\
        & attribute identification &  Specifies attributes of objects (e.g., color, shape, size) \\
        \midrule
        \multirow{5}{*}{\rotatebox{90}{\bf Reasoning}}
        & object measuring &  Measures object dimensions or distances in the image \\
        & arithmetic calculation &  Comprises operations such as Addition, Subtraction, Multiplication and Division \\
        & counting &  Handles counting queries such as 'how many...' \\
        & comparison &  Handles comparison queries \\
        & logical operations &  Handles logical operations such as AND (intersection) and OR (union) on sets \\
        \midrule
        \multirow{6}{*}{\rotatebox{90}{\bf Tools}}
        & QR code scanning &  Scans and decodes QR codes from the image \\
        & OCR &  Optical Character Recognition — extracts text and their locations in the image \\
        & GUI &  Handles graphical user interface-based tasks \\
        & knowledge search &  Searches for external knowledge beyond the image itself \\
        & using external API &  Uses an external API such as image enhancement and reading metadata. \\
        & knowledge retrieval &  Retrieve the necessary information in the metadata of images \\
        \bottomrule
    \end{tabular}
    }

    \label{tab:skill_taxonomy}
\end{table*}

\subsection{Data Annotation}

All questions in \name~  were designed and annotated by the authors. Each question was carefully crafted and reviewed to ensure clarity, relevance, and alignment with a corresponding image. An exact-match answer was provided for each question, establishing a clear ground truth.

To ensure the quality, solvability, and objectivity of the dataset, the annotation process was divided into three distinct phases. In the first phase, one author was responsible for collecting the images and constructing the initial set of questions. In the second phase, five other authors independently reviewed the questions and provided feedback to refine and clarify them where necessary. In the third phase, to promote annotation consistency and minimize bias, an internal tool was developed to randomly assign each question to at least two reviewers. Reviewers answered the questions without any additional context beyond the image and the question text. 

Any question that could not be reliably answered by the reviewers without additional input from the original author was removed from the dataset. This three-phase process ensured that the final dataset includes only questions that are independently answerable and clearly grounded in the accompanying visual content.

\begin{figure}[t]
    \centering
    \includegraphics[width=0.99\linewidth]{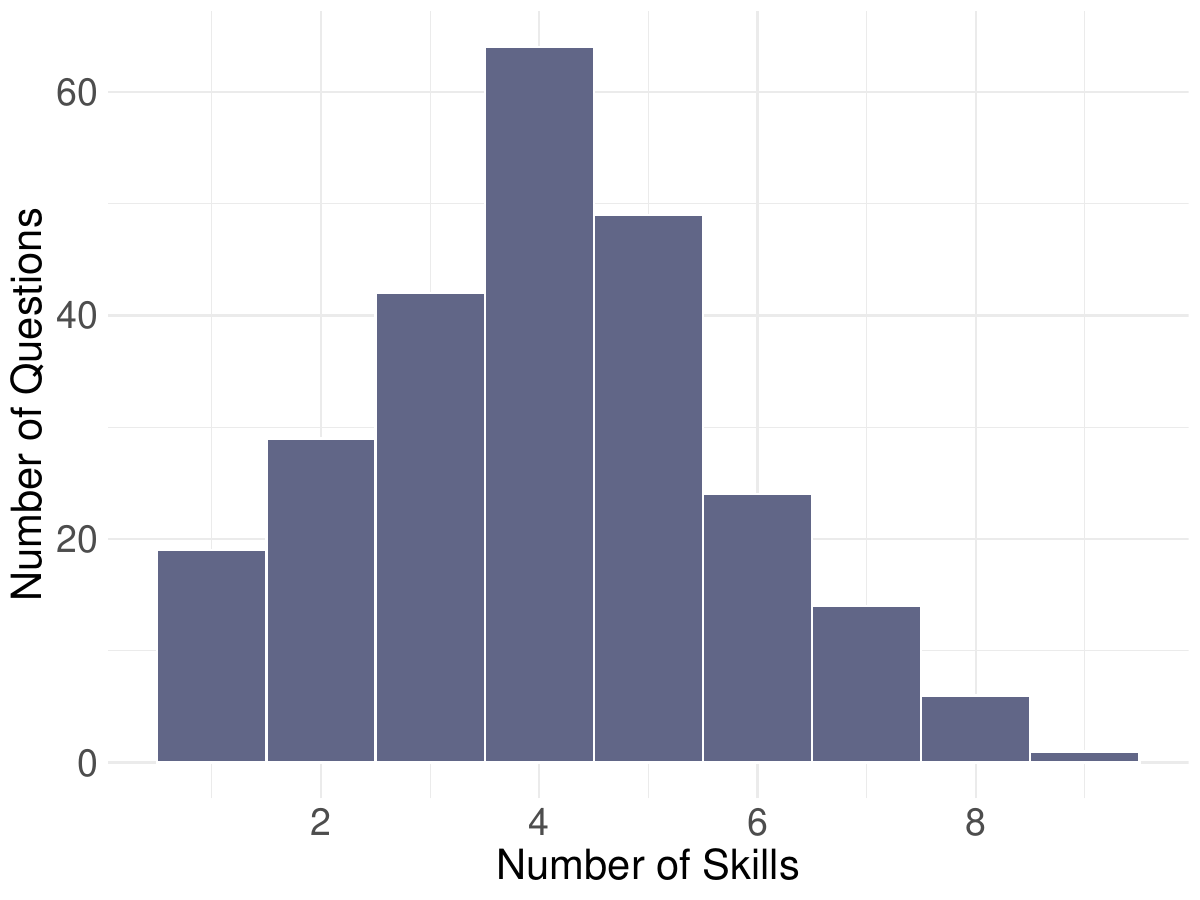}
    \caption{Number of unique skills used per question.}
    \label{fig:reasoning-step}
\end{figure}

\subsection{ Answer Format Standardization}

We explicitly specify the format of the expected answer for each question. This includes defining whether the answer should be a \textit{yes/no response}, a \textit{multiple-choice selection}, or a \textit{short exact-match answer}. For some questions, multiple answers are allowed. Acceptable answers are separated by semicolons. The exact output format instruction is provided in Appendix \ref{app:answer_format}. By standardizing the answer format, we facilitate straightforward evaluation, enabling seamless testing of both exact-match answers and multiple-choice responses. This design ensures that the dataset is both practical and robust, catering to diverse testing and benchmarking needs while maintaining ease of implementation.

It is important to note that, in order to conform to an exact-match evaluation format, questions were transformed into constrained response types such as multiple choice, yes/no, single numerical values, or short text answers limited to a few words. While this approach facilitates consistent evaluation and benchmarking across models, it may also increase the likelihood of correct responses, as it narrows the range of possible outputs.

Preliminary experiments using free-form, open-ended answers reveal significantly higher failure rates, suggesting that exact-match formats may overestimate model performance by simplifying the response space. This is a trade-off between evaluation consistency and the complexity of real-world language understanding.

\begin{figure}[t]
    \centering
    \includegraphics[width=0.8\linewidth]{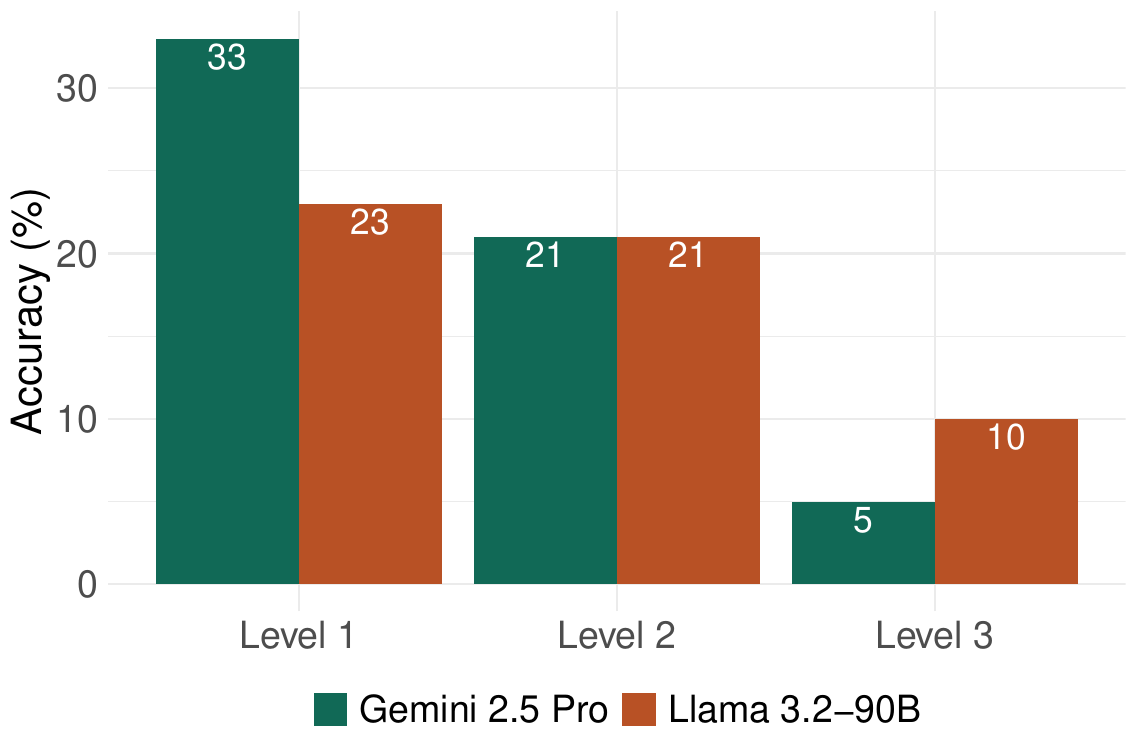}
    \caption{\textbf{Model performance degradation across difficulty levels.} Results show both Gemini 2.5 Pro and Llama 3.2-70B exhibit declining accuracy as task difficulty increases, illustrating current limitations in complex visual reasoning. }
    \label{fig:performance-by-difficulty-level}
\end{figure}

\subsection{Difficulty Level}

Following \cite{mialon2023gaia}, we categorize questions into three levels of increasing difficulty based on the size of unique skills needed to answer the question. Figure \ref{fig:reasoning-step} shows the a number of unique skills used per question.

We broadly define difficulty levels as follows:
\begin{itemize}
    \item Level 1: Typically involves no more than 2 unique skills and at most 1 external tool.
    \item Level 2: Involves a greater number of skills - generally between 3 and 5 skills, typically includes a combination of two tools.
    \item Level 3: Designed for an ideal general-purpose assistant, these questions may require arbitrarily long sequences of actions, unrestricted use of tools, and general access to the whole Internet.
\end{itemize}



Figure \ref{fig:performance-by-difficulty-level} shows the performance of Gemini Pro and Llama 3.2 90B grouped by difficulty level. Overall, these VLMs' performance deteriorates, highlighting the persistent gap between current AI capabilities and human-like reasoning.

\begin{figure}[t]
    \centering
    \includegraphics[width=\linewidth]{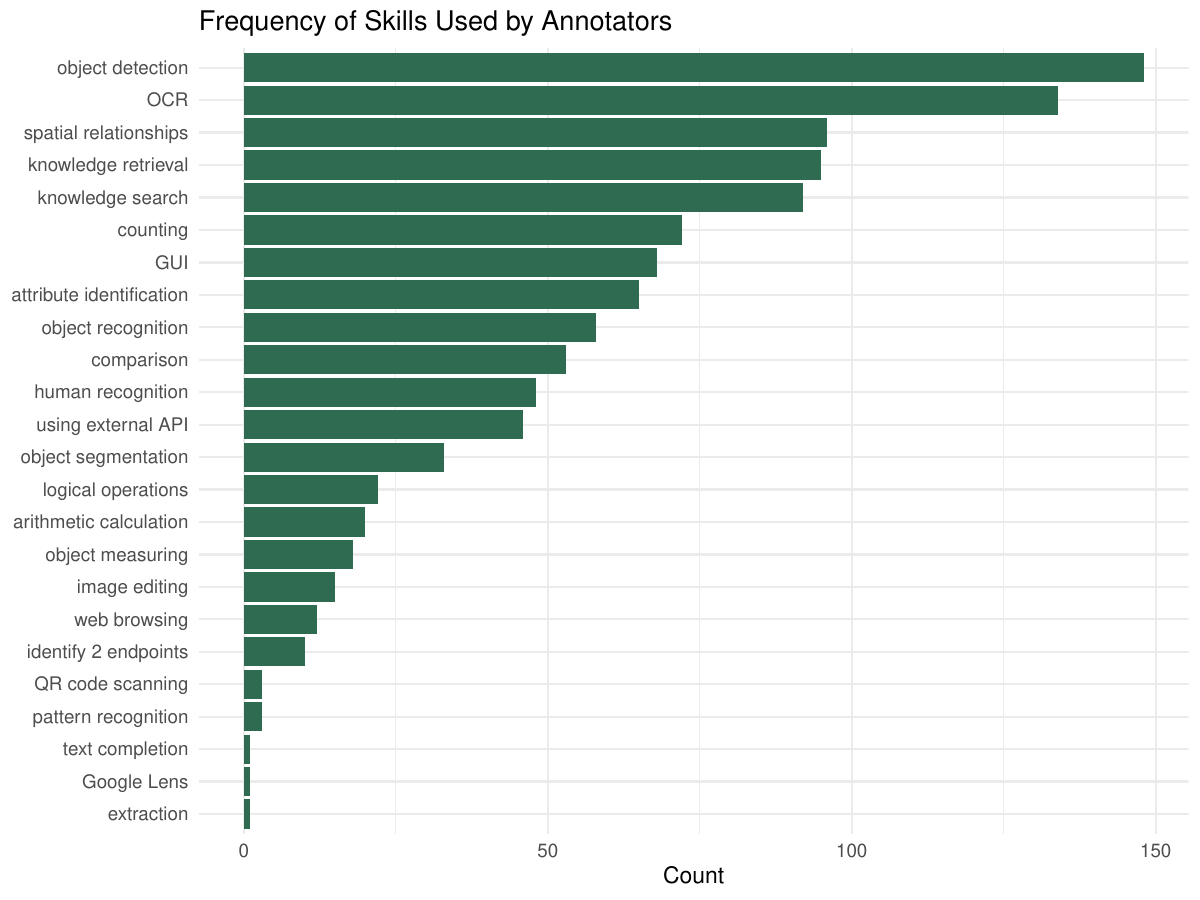}
    \caption{Distribution of skills annotated by human. Noted that annotators are allowed to add additional skills or tools that they used (e.g. Google Lens)}.
    \label{fig:skill_by_annotators}
\end{figure}

We also displays the distribution of skills needed that were generated by annotators in Figure \ref{fig:skill_by_annotators} . Among visual skills, object detection, OCR, and spatial reasoning are most common. In reasoning skills, counting and knowledge retrieval dominate. For tool-use skills, knowledge search and GUI interaction are most frequent.

\section{Experiments}

In this session, we assess the capabilities of three powerful methodological approaches on OWLViz: Vanilla VLMs, Tool-Calling Agents and GUI Agents. This comparative analysis allows us to identify and examine the key challenges presented by our dataset. 

\subsection{Evaluation Metrics}
We employ two metrics to evaluate model performance: \emph{Exact Match} ({\bf EM}) and \emph{LLM-based Match} ({\bf LM}) \cite{zheng2023judging}. The EM metric requires the model's output to be identical to the ground truth answer, allowing for minor variations in capitalization and whitespace. In practice, some models still fail to follow output format instructions, leading to false negative errors. To allow for more flexible evaluation, we use LM, where GPT-4o acts as a judge to determine semantic equivalence between the predicted and ground truth answers (See Appendix \ref{app:llm_match}).

\subsection{Vanilla VLMs}

\paragraph{Models.} We evaluate a wide set of open-source VLM models: DeepSeek-VL \cite{wu2024deepseek}, Qwen-VL \cite{yang2024qwen2}, InternVL \cite{zhang-zhou-2022-temporal}, LLaVa \cite{liu2023llava}, and Molmo \cite{deitke2024molmo} as well as proprietary models, i.e., Anthropic's Claude, OpenAI's GPT4V/o, and Google's Gemini.

\begin{table}[t]
\caption{Model performance breakdown into 3 groups ordered by EM performance. The best and second-best of each group are {\bf bolded} and \underline{underlined}, respectively.}
    \resizebox{1.0\linewidth}{!}{
    \begin{tabular}{clrr}
        \toprule
        & \bf Model & \bf EM & \bf LM \\
        \midrule
        & \bf Human                     & \multicolumn{2}{c}{\bf 69.21} \\
        \midrule
        \multirow{11}{*}{\rotatebox{90}{\bf Small Open Source}}
        & DeepSeek-VL2-small (2.8B)     & 11.16 & 12.75 \\
        & DeepSeek-VL2 (4.5B active)    & 11.16 & 14.34 \\
        & Qwen2-VL-7B-Instruct          & 12.75 & 17.93 \\
        & Qwen2.5-VL-7B-Instruct        & 13.94 & 19.52 \\
        & InternVL3-8B                  & 14.34 & \underline{21.12} \\
        & LLaVa-v1.6-mistral-7B         & 14.74 & 15.54 \\
        & Llama-3.2-11B-Vision-Instruct & 14.74 & {\bf 25.10} \\
        & InternVL2.5-8B                & 14.74 & 18.73 \\
        & LLaVa-v1.5-13B                & 16.33 & 16.33 \\
        & Molmo-7B-D-0924               & \underline{17.13} & 20.32 \\
        & LLaVa-v1.5-7B                & {\bf 18.33} & 19.92 \\

        \midrule
        \multirow{9}{*}{\rotatebox{90}{\bf Large Open Source}}
        & Qwen2.5-VL-32B-Instruct       & 2.79  & \underline{25.90} \\
        & InternVL2.5-38B               & 13.94 & 19.52 \\
        & InternVL3-78B                 & 15.54 & 20.72 \\
        & Molmo-72B-0924                & 15.94 & 22.71 \\
        & InternVL2.5-78B               & 15.94 & 21.91 \\
        & InternVL3-38B                 & 16.73 & 23.11 \\
        & Qwen2-VL-72B-Instruct         & 19.92 & 25.90 \\
        & Qwen2.5-VL-72B-Instruct       & \underline{20.32} & {\bf 26.29} \\
        & Llama-3.2-90B-Vision-Instruct & {\bf 20.72} & 24.70 \\
        \midrule
        \multirow{7}{*}{\rotatebox{90}{\bf Proprietary}}
        & Claude-3-5-sonnet-20241022    & 11.55 & 19.92 \\
        & GPT-4V                        & 14.34 & 20.00 \\
        & Gemini-2.5-Flash              & 15.54 & \underline{25.50} \\
        & GPT-4o (\texttt{2024-11-20})  & 16.33 & 19.52 \\
        & Gemini-1.5-Pro                & \underline{19.52} & 21.91 \\
        & Gemini-2.0-Flash              & {\bf 21.51} & 24.30 \\
        & Gemini-2.5-Pro                & {\bf 21.51} & {\bf 27.09} \\
        \bottomrule
    \end{tabular}
}

\label{tab:eval-vlm}
\end{table}

\paragraph{Results.}
Table \ref{tab:eval-vlm} presents our evaluation of VLMs across three categories: small open-source, large open-source, and proprietary models. The results demonstrate that \name~poses significant challenges even for state-of-the-art VLMs. The best-performing models, Gemini-2.5-pro-preview and Gemini-2.0-flash, achieve only 21.51\% EM accuracy, indicating substantial room for improvement on this benchmark. This relatively low performance across all models suggests that our dataset effectively probes the limitations of current VLMs in handling complex visual reasoning tasks when not equipped with tools.

Within the open-source category, larger models generally perform better than their smaller counterparts, but the improvement is modest. For instance, Llama-3.2-11B-Vision-Instruct tops the small model leaderboard at 25.10\% LM, its larger variant Llama-3.2-90B-Vision-Instruct yields a lower LM score (24.70\%). This suggests that simply scaling up model size may not be sufficient to address the challenging nature of our benchmark. Among proprietary models, while Gemini-2.0-Flash sets the current state-of-the-art, its performance (21.51\% EM, 27.09\% LM) still falls significantly short of human-level understanding, highlighting the substantial gap between current AI capabilities and human visual reasoning abilities. Notably, the consistent gap between EM and LM scores across all models indicates that even when models grasp the correct concept, they often struggle to express it in the exact required format.

\subsection{Tool-Calling Agents}

\paragraph{Models.}
Tool-Calling Agents extend traditional VLMs by integrating external tools and action capabilities to solve complex visual reasoning tasks. Our evaluation examines six representative systems: LLaVa-Plus \cite{liu2025llavaplus}, which enhances LLaVA \cite{liu2023llava} with pre-trained vision tools for improved reasoning; ViperGPT \cite{suris2023vipergpt}, which uses GPT-4o to generate Python code orchestrating multiple vision models; GPT4Tools \cite{yang2023gpt4tools}, built on Vicuna with instruction tuning for visual tool control; HYDRA \cite{jalaian2025hydra}, employing deep reinforcement learning to fine-tune LLMs for dynamic visual reasoning; HF Agent includes predefined tools for visual tasks and web browsing; and DynaSaur \cite{nguyen2024dynasaur}, supporting real-time generation and composition of actions with capabilities for continual learning through action storage and reuse. 

\begin{table}[t]
    \caption{Performance of Agentic models with tool-uses. The best and second-best of each group are {\bf bolded} and \underline{underlined}, respectively.}
    \resizebox{\linewidth}{!}{
    \begin{tabular}{l|c|c|rr}
        \toprule
        \bf Model & \bf MLLM & \bf EM & \bf LM \\
        \midrule
        LLaVa-Plus  &	gpt-4o-2024-11-20 & 0.00 & 2.50 \\
        ViperGPT    &   gpt-4o-2024-11-20 & 7.56 & 12.35 \\
        GPT4Tools   &   vicuna-7b-v1.5    & 11.15 & 14.34 \\
        HYDRA       & 	gpt-4o-2024-11-20 & 10.75 & 12.35\\
        HF Agent    &	gpt-4o-2024-11-20 & \bf 18.32 & \underline{24.08}\\
        DynaSaur    &   gpt-4o-2024-11-20 & \underline{16.23} & \bf 26.67\\
        \bottomrule
    \end{tabular}
    }
    \label{tab:eval-tool-use}
\end{table}

\paragraph{Results.} Table \ref{tab:eval-tool-use} presents the performance comparison of various large language models (LLM) agents equipped with tool-use capability. 
Among the models, HF Agent achieves the highest EM score (18.32\%), indicating superior accuracy in exact task execution, followed closely by DynaSaur (16.23\%). DynaSaur leads in LM score (26.67\%), suggesting strong overall language understanding and generation capabilities. In contrast, LLava-Plus performs poorly across both metrics, particularly with an EM score of 0.00, highlighting limitations in task precision.

\paragraph{Tool Exploration Incentives.}

\begin{figure}[t]
    \centering
    \includegraphics[width=\linewidth]{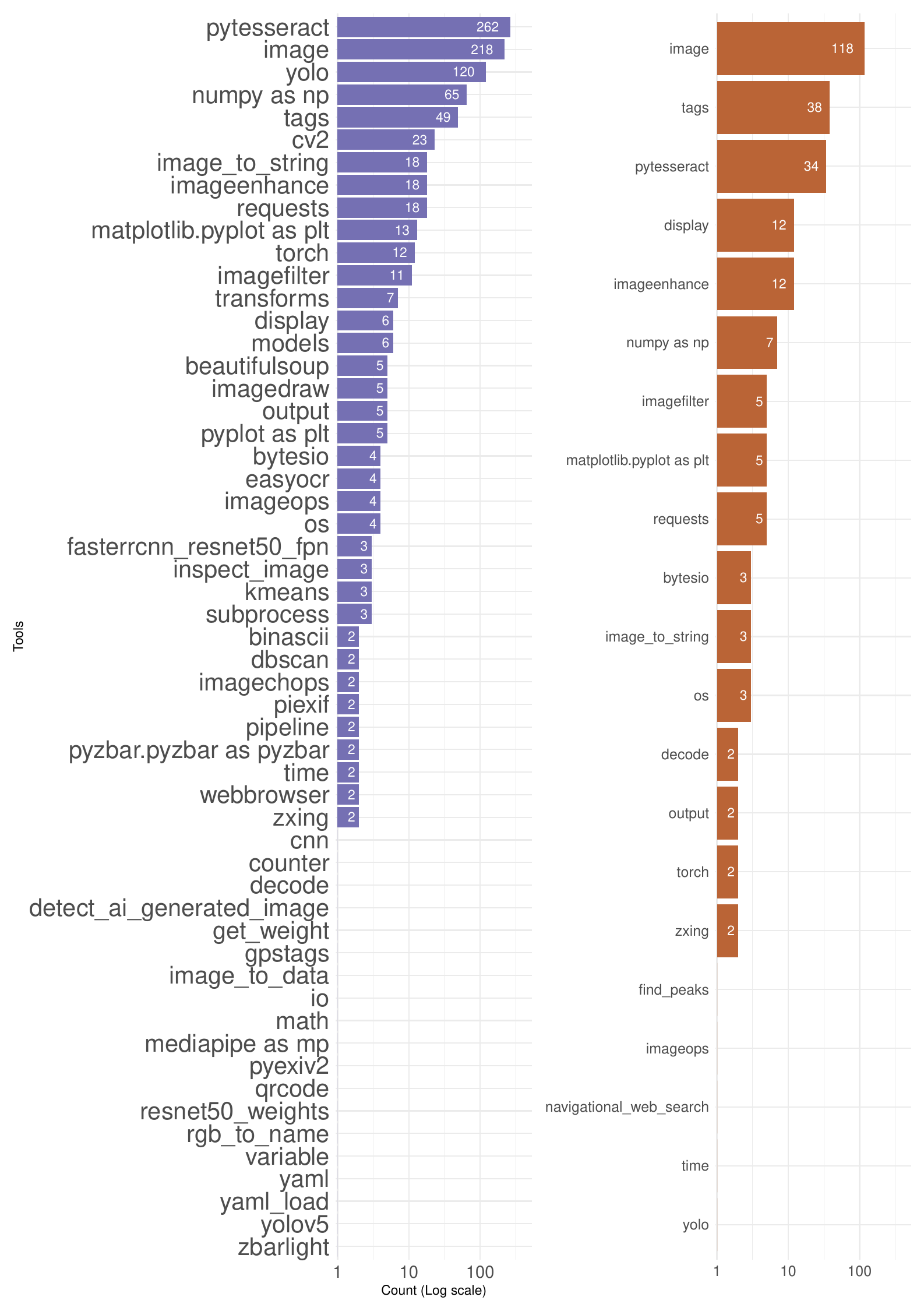}
    \caption{Comparison of libraries and ML models used by DynaSaur with (left) and without incentive (right).}
    \label{fig:with_without_incentive}
\end{figure}

Our analysis reveals that the original tool-calling agent baselines demonstrate insufficient motivation to utilize external models or tools for visual question-answering tasks, instead defaulting to generating code comments for reasoning before directly producing answers. To mitigate this limitation, we implemented an explicit instruction requiring agents to employ at least one external tool for answer verification prior to submission. We subsequently conducted a comprehensive analysis of all libraries and machine learning models generated by DynaSaur across 100 sample instances. Figure \ref{fig:with_without_incentive} illustrates the comparative frequency of tool utilization between models with and without these incentives. This strategic modification yielded a 2\% improvement in Exact Match (EM) score. For detailed results, readers may refer to Table \ref{tab:app-tool-exploration-incentive} in the appendix.

\begin{table}[!h]
    \centering
    \caption{Performance of LLM with tool-use}
    \begin{tabular}{lrr}
        \toprule
        \bf Model \hspace{2cm} & \bf EM \\
        \midrule
        DynaSaur (w/o incentive) & 9.0\\
        DynaSaur (w/ incentive) & 11.0 \\
        \bottomrule
    \end{tabular}
    
    \label{tab:app-tool-exploration-incentive}
\end{table}

In the no-incentive scenario (Figure \ref{fig:with_without_incentive}), tool usage is more concentrated and limited in diversity, with image (118), tags (38), and pytesseract (34) being the most frequently used tools. Most other tools appear only a handful of times, reflecting a poor coverage of the tools being used. In contrast, models with incentives show both a significant increase in tool usage and a broader variety of tools. pytesseract leads with 262 mentions, followed by image (218), and YOLO (120). Additionally, a wide array of tools such as cv2, torch, transforms, and various deep learning components (e.g., fasterrcnn\_resnet50\_fpn, cnn) appear in the incentivized setting. This indicates that the presence of incentives not only increases effort but also encourages a broader adoption of more sophisticated, task-specific tools.

\begin{table}[t]
\caption{The performance in EM and LM of GUI Agents on OWLViz with the average number of mouse actions.}
\resizebox{\linewidth}{!}{
    \begin{tabular}{l|rr|ccc}
        \toprule
        \bf Model & \bf EM & \bf LM & \bf Click & \bf Hover & \bf Scroll \\
        \midrule
        UI-TARS  & 0.00 & 12.31 & 0.91 & 0.51 & 0.68\\
        ShowUI  & 0.00 & 12.80 & 0.97 & 0.19 & 0.10\\
        \bottomrule
    \end{tabular}
}
\label{tab:eval-gui-tool-use}
\end{table}

\subsection{GUI Agents}
\paragraph{Models.}
Our evaluation includes two GUI interaction baselines combining GPT-4o (2024-11-20) with specialized action determination components. UItars \cite{qin2025ui} pairs GPT-4o's reasoning capabilities with UI-TARS-2B-SFT for action determination in graphical interface tasks, enabling structured human-computer interactions. Similarly, ShowUI \cite{lin2024showui} employs the same GPT-4o model for reasoning but integrates ShowUI-2B for GUI action determination, facilitating efficient multimodal instruction-following in computer interfaces.

\begin{figure*}[!h]
    \centering
    \begin{minipage}[t]{0.35\textwidth}
        \begin{tcolorbox}[height=7cm, colback=cyan!5!white, colframe=cyan!75!white, title=Question \& Answer]
            What is the name of the shop that is located across the street from the lot for sale in this photo? Provide an answer in fewer than 3 words\\
            
            {\bf Any of the following answers are acceptable:} \\
            Wheat Bay; \\
            Uniquely Chengdu.
        \end{tcolorbox}
    \end{minipage}
    \hfill
    \begin{minipage}[t]{0.63\textwidth}
        \includegraphics[height=7cm]{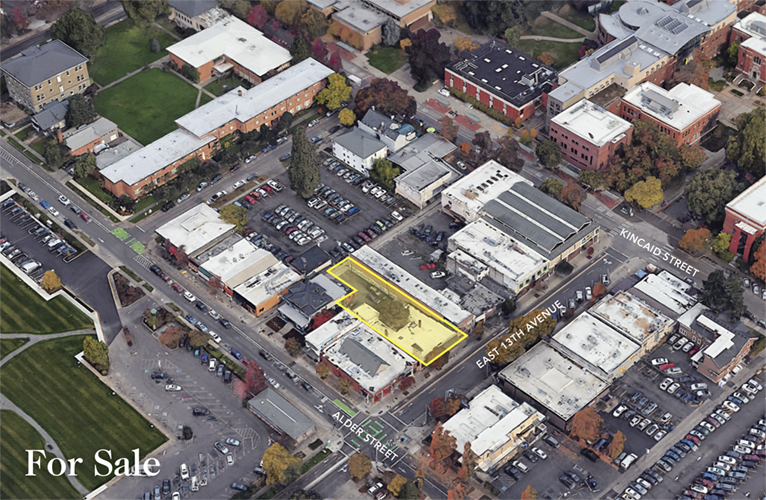}
    \end{minipage}

    \vspace{1em}

    \begin{minipage}[t]{0.37\textwidth}
        \begin{tcolorbox}[height=2.5cm, colback=blue!5!white, colframe=blue!75!white, title=Gemini]
            \it ...Identify the shop across Alder Street from the for-sale lot. \\
            Not identifiable
        \end{tcolorbox}
    \end{minipage}
    \hfill
    \begin{minipage}[t]{0.36\textwidth}
        \begin{tcolorbox}[height=2.5cm, colback=green!5!white, colframe=green!75!black, title=DynaSaur]
                \it The name of the shop across the street is visible in the image. It is "Starbucks Coffee".
        \end{tcolorbox}
        
    \end{minipage}
    \hfill
    \begin{minipage}[t]{0.24\textwidth}
        \begin{tcolorbox}[height=2.5cm, colback=red!5!white, colframe=red!75!white, title=ShowUI]
            \it 
            I need to...\\
            Action: Scroll \\
            No answer
        \end{tcolorbox}
    \end{minipage}

    \caption{\textbf{Qualitative results comparing different model capabilities on OWLViz.} Results demonstrate varying capabilities across model types: Gemini (vanilla VLM) fails to identify the target, DynaSaur (tool-calling agent) produces an incorrect answer despite external search capabilities, and ShowUI (GUI agent) provides no answer. }
\label{fig:quanlitative_result}
\end{figure*}

\paragraph{Results.}
Table \ref{tab:eval-gui-tool-use} presents the performance of two GUI agents—UI-TARS and ShowUI—evaluated in task accuracy and the average number of mouse actions (Click, Hover, Scroll). Both agents fail to achieve any correct task as indicated by their EM scores of 0.00, highlighting a complete lack of task-following capabilities in output format. While LM scores are modest, with ShowUI slightly better than UI-TARS (12.80\% vs. 12.31\%).

Notably, the total number of actions performed by these agents is extremely low, reflecting minimal engagement with the interface. UI-TARS averaged fewer than 1 click (0.91), hover (0.51), and scroll (0.68) actions per task, while ShowUI executed slightly more clicks (0.97) but significantly fewer hovers (0.19) and scrolls (0.10). These numbers indicate that these GUI agents barely interact with the user interface. This somehow explains the low performance of the GUI agents, even when the model is allowed to explore freely.

\subsection{Comparison between orchestrations}

In our evaluation of model performance across vision-language models (VLMs), agent-based systems, and GUI agents, we observe distinct capability profiles across these paradigms. Vision-language models, particularly proprietary systems such as Gemini-2.5-Pro, demonstrate the strongest overall performance, achieving the highest exact match (EM) and LLM match (LM) scores. Agent-based systems that incorporate tool use, such as HF Agent and DynaSaur powered by GPT-4o, show a marked improvement in EM over their base models (e.g., HF Agent achieves 18.32 EM vs. 16.33 for standalone GPT-4o), suggesting that tool augmentation enables more effective task execution and grounded reasoning. However, GUI agents such as UI-TARS and ShowUI exhibit significant limitations, with EM scores remaining at 0.00 and LM scores peaking at only 12.80\%. Their low interaction metrics across click, hover, and scroll actions further highlight their current inability to perform even basic interface-driven tasks reliably. 

\subsection{Qualitative Example}

We present a qualitative example that demonstrates the varying capabilities of different model types (see Figure \ref{fig:quanlitative_result}). Please refer to Section \ref{app:detailed_examples} in the supplementary materials for detailed output.

\begin{itemize}
    \item {\bf Gemini}, representing vanilla VLMs, attempts to locate the "For Sale" lot but fails to identify the street name and the requested business, resulting in a conclusion "Not identifiable". 
    \item {\bf DynaSaur}, the tool-calling agent, leverages OCR and Search to identify the correct lot address ("821-825 E 13th Ave, Eugene, OR") and employs external search tools to find businesses in the vicinity. Through progressive refinement of search queries and cross-referencing, DynaSaur identifies "Starbucks Coffee" as the shop across from the lot. However, this is an incorrect answer, highlighting challenges in accurate spatial reasoning and external knowledge integration (e.g., maps with street view). 
    \item {\bf ShowUI}, representing GUI agents, demonstrates the limitations of restricted action capabilities. With only two available actions, the model attempts to scroll but ultimately fails to produce an answer. This illustrates how constraints on interaction modalities can significantly impact performance on complex visual reasoning tasks. 
\end{itemize}

\section{Conclusion}

We introduced \name, a challenging benchmark for evaluating AI models' visual understanding, reasoning, and tool use. By integrating real-world tasks requiring image comprehension, metadata extraction, web exploration, and external tool use, \name~ highlights the limitations of current VLMs, tool-use agents, and GUI agents, which fall short of human performance. Our results show that AI struggles with multi-step reasoning and practical tool integration, underscoring the need for further advancements. 





\bibliography{custom,anthology.2022.2024,anthology.2024.2026}

\begin{thebibliography}{46}
\providecommand{\natexlab}[1]{#1}
\providecommand{\url}[1]{\texttt{#1}}
\expandafter\ifx\csname urlstyle\endcsname\relax
  \providecommand{\doi}[1]{doi: #1}\else
  \providecommand{\doi}{doi: \begingroup \urlstyle{rm}\Url}\fi

\bibitem[Antol et~al.(2015)Antol, Agrawal, Lu, Mitchell, Batra, Zitnick, and Parikh]{nntol2015vqa}
Antol, S., Agrawal, A., Lu, J., Mitchell, M., Batra, D., Zitnick, C.~L., and Parikh, D.
\newblock Vqa: Visual question answering.
\newblock In \emph{Proceedings of the IEEE International Conference on Computer Vision (ICCV)}, December 2015.

\bibitem[Bitton et~al.(2023)Bitton, Bansal, Hessel, Shao, Zhu, Awadalla, Gardner, Taori, and Schmidt]{bitton2023visit}
Bitton, Y., Bansal, H., Hessel, J., Shao, R., Zhu, W., Awadalla, A., Gardner, J., Taori, R., and Schmidt, L.
\newblock Visit-bench: A benchmark for vision-language instruction following inspired by real-world use.
\newblock \emph{arXiv preprint arXiv:2308.06595}, 2023.

\bibitem[Brohan et~al.(2023)Brohan, Brown, Carbajal, Chebotar, Chen, Choromanski, Ding, Driess, Dubey, Finn, et~al.]{brohan2023rt}
Brohan, A., Brown, N., Carbajal, J., Chebotar, Y., Chen, X., Choromanski, K., Ding, T., Driess, D., Dubey, A., Finn, C., et~al.
\newblock Rt-2: Vision-language-action models transfer web knowledge to robotic control.
\newblock \emph{arXiv preprint arXiv:2307.15818}, 2023.

\bibitem[Chen et~al.(2024)Chen, Wang, Cao, Liu, Gao, Cui, Zhu, Ye, Tian, Liu, et~al.]{chen2024expanding}
Chen, Z., Wang, W., Cao, Y., Liu, Y., Gao, Z., Cui, E., Zhu, J., Ye, S., Tian, H., Liu, Z., et~al.
\newblock Expanding performance boundaries of open-source multimodal models with model, data, and test-time scaling.
\newblock \emph{arXiv preprint arXiv:2412.05271}, 2024.

\bibitem[Dang et~al.(2024)Dang, Le, Le, Phuong, and Tran]{dang2024sadl}
Dang, L.~H., Le, T.~M., Le, V., Phuong, T.~M., and Tran, T.
\newblock Sadl: An effective in-context learning method for compositional visual qa.
\newblock \emph{CoRR}, 2024.

\bibitem[Deitke et~al.(2024)Deitke, Clark, Lee, Tripathi, Yang, Park, Salehi, Muennighoff, Lo, Soldaini, et~al.]{deitke2024molmo}
Deitke, M., Clark, C., Lee, S., Tripathi, R., Yang, Y., Park, J.~S., Salehi, M., Muennighoff, N., Lo, K., Soldaini, L., et~al.
\newblock Molmo and pixmo: Open weights and open data for state-of-the-art multimodal models.
\newblock \emph{arXiv preprint arXiv:2409.17146}, 2024.

\bibitem[Dubey et~al.(2024)Dubey, Jauhri, Pandey, Kadian, Al-Dahle, Letman, Mathur, Schelten, Yang, Fan, et~al.]{dubey2024llama}
Dubey, A., Jauhri, A., Pandey, A., Kadian, A., Al-Dahle, A., Letman, A., Mathur, A., Schelten, A., Yang, A., Fan, A., et~al.
\newblock The llama 3 herd of models.
\newblock \emph{arXiv preprint arXiv:2407.21783}, 2024.

\bibitem[Gao et~al.(2023)Gao, Wang, Shan, and Chen]{gao2023cric}
Gao, D., Wang, R., Shan, S., and Chen, X.
\newblock Cric: A vqa dataset for compositional reasoning on vision and commonsense.
\newblock \emph{IEEE Transactions on Pattern Analysis and Machine Intelligence}, 45\penalty0 (5):\penalty0 5561--5578, 2023.
\newblock \doi{10.1109/TPAMI.2022.3210780}.

\bibitem[Glazer et~al.(2024)Glazer, Erdil, Besiroglu, Chicharro, Chen, Gunning, Olsson, Denain, Ho, de~Oliveira~Santos, J{\"{a}}rviniemi, Barnett, Sandler, Vrzala, Sevilla, Ren, Pratt, Levine, Barkley, Stewart, Grechuk, Grechuk, Enugandla, and Wildon]{DBLP:journals/corr/abs-2411-04872}
Glazer, E., Erdil, E., Besiroglu, T., Chicharro, D., Chen, E., Gunning, A., Olsson, C.~F., Denain, J., Ho, A., de~Oliveira~Santos, E., J{\"{a}}rviniemi, O., Barnett, M., Sandler, R., Vrzala, M., Sevilla, J., Ren, Q., Pratt, E., Levine, L., Barkley, G., Stewart, N., Grechuk, B., Grechuk, T., Enugandla, S.~V., and Wildon, M.
\newblock Frontiermath: {A} benchmark for evaluating advanced mathematical reasoning in {AI}.
\newblock \emph{CoRR}, abs/2411.04872, 2024.
\newblock \doi{10.48550/ARXIV.2411.04872}.
\newblock URL \url{https://doi.org/10.48550/arXiv.2411.04872}.

\bibitem[Goyal et~al.(2017)Goyal, Khot, Summers{-}Stay, Batra, and Parikh]{balanced_vqa_v2}
Goyal, Y., Khot, T., Summers{-}Stay, D., Batra, D., and Parikh, D.
\newblock Making the {V} in {VQA} matter: Elevating the role of image understanding in {V}isual {Q}uestion {A}nswering.
\newblock In \emph{Conference on Computer Vision and Pattern Recognition (CVPR)}, 2017.

\bibitem[Hendrycks et~al.(2020)Hendrycks, Burns, Basart, Zou, Mazeika, Song, and Steinhardt]{hendrycks2020measuring}
Hendrycks, D., Burns, C., Basart, S., Zou, A., Mazeika, M., Song, D., and Steinhardt, J.
\newblock Measuring massive multitask language understanding.
\newblock \emph{arXiv preprint arXiv:2009.03300}, 2020.

\bibitem[Hu et~al.(2024{\natexlab{a}})Hu, Stretcu, Lu, Viswanathan, Hata, Luo, Krishna, and Fuxman]{DBLP:conf/cvpr/HuSLVHLKF24}
Hu, Y., Stretcu, O., Lu, C., Viswanathan, K., Hata, K., Luo, E., Krishna, R., and Fuxman, A.
\newblock Visual program distillation: Distilling tools and programmatic reasoning into vision-language models.
\newblock In \emph{{IEEE/CVF} Conference on Computer Vision and Pattern Recognition, {CVPR} 2024, Seattle, WA, USA, June 16-22, 2024}, pp.\  9590--9601. {IEEE}, 2024{\natexlab{a}}.
\newblock \doi{10.1109/CVPR52733.2024.00916}.
\newblock URL \url{https://doi.org/10.1109/CVPR52733.2024.00916}.

\bibitem[Hu et~al.(2024{\natexlab{b}})Hu, Ren, Li, and Yin]{hu-etal-2024-viva}
Hu, Z., Ren, Y., Li, J., and Yin, Y.
\newblock {VIVA}: A benchmark for vision-grounded decision-making with human values.
\newblock In Al-Onaizan, Y., Bansal, M., and Chen, Y.-N. (eds.), \emph{Proceedings of the 2024 Conference on Empirical Methods in Natural Language Processing}, pp.\  2294--2311, Miami, Florida, USA, November 2024{\natexlab{b}}. Association for Computational Linguistics.
\newblock \doi{10.18653/v1/2024.emnlp-main.137}.
\newblock URL \url{https://aclanthology.org/2024.emnlp-main.137/}.

\bibitem[Hudson \& Manning(2019)Hudson and Manning]{hudson2019gqa}
Hudson, D.~A. and Manning, C.~D.
\newblock Gqa: A new dataset for real-world visual reasoning and compositional question answering.
\newblock In \emph{Proceedings of the IEEE/CVF Conference on Computer Vision and Pattern Recognition (CVPR)}, June 2019.

\bibitem[Jalaian et~al.(2025)Jalaian, Bastian, et~al.]{jalaian2025hydra}
Jalaian, B., Bastian, N.~D., et~al.
\newblock Hydra: An agentic reasoning approach for enhancing adversarial robustness and mitigating hallucinations in vision-language models.
\newblock \emph{arXiv preprint arXiv:2504.14395}, 2025.

\bibitem[Jimenez et~al.(2024)Jimenez, Yang, Wettig, Yao, Pei, Press, and Narasimhan]{DBLP:conf/iclr/JimenezYWYPPN24}
Jimenez, C.~E., Yang, J., Wettig, A., Yao, S., Pei, K., Press, O., and Narasimhan, K.~R.
\newblock Swe-bench: Can language models resolve real-world github issues?
\newblock In \emph{The Twelfth International Conference on Learning Representations, {ICLR} 2024, Vienna, Austria, May 7-11, 2024}. OpenReview.net, 2024.
\newblock URL \url{https://openreview.net/forum?id=VTF8yNQM66}.

\bibitem[Jing et~al.(2024)Jing, Li, Chen, and Du]{jing-etal-2024-faithscore}
Jing, L., Li, R., Chen, Y., and Du, X.
\newblock {F}aith{S}core: Fine-grained evaluations of hallucinations in large vision-language models.
\newblock In Al-Onaizan, Y., Bansal, M., and Chen, Y.-N. (eds.), \emph{Findings of the Association for Computational Linguistics: EMNLP 2024}, pp.\  5042--5063, Miami, Florida, USA, November 2024. Association for Computational Linguistics.
\newblock \doi{10.18653/v1/2024.findings-emnlp.290}.
\newblock URL \url{https://aclanthology.org/2024.findings-emnlp.290/}.

\bibitem[Johnson et~al.(2017)Johnson, Hariharan, Van Der~Maaten, Fei-Fei, Lawrence~Zitnick, and Girshick]{johnson2017clevr}
Johnson, J., Hariharan, B., Van Der~Maaten, L., Fei-Fei, L., Lawrence~Zitnick, C., and Girshick, R.
\newblock Clevr: A diagnostic dataset for compositional language and elementary visual reasoning.
\newblock In \emph{Proceedings of the IEEE conference on computer vision and pattern recognition}, pp.\  2901--2910, 2017.

\bibitem[Kelly et~al.(2024)Kelly, Hu, Yang, Tian, Yang, Yang, Huang, Li, Hu, and Zou]{DBLP:journals/corr/abs-2403-09027}
Kelly, C., Hu, L., Yang, B., Tian, Y., Yang, D., Yang, C., Huang, Z., Li, Z., Hu, J., and Zou, Y.
\newblock Visiongpt: Vision-language understanding agent using generalized multimodal framework.
\newblock \emph{CoRR}, abs/2403.09027, 2024.
\newblock \doi{10.48550/ARXIV.2403.09027}.
\newblock URL \url{https://doi.org/10.48550/arXiv.2403.09027}.

\bibitem[Le et~al.(2024)Le, Dang, Le, Tran, and Le]{le2024progressive}
Le, Q.-H., Dang, L.~H., Le, N., Tran, T., and Le, T.~M.
\newblock Progressive multi-granular alignments for grounded reasoning in large vision-language models.
\newblock \emph{arXiv preprint arXiv:2412.08125}, 2024.

\bibitem[Li et~al.(2023)Li, Zhao, Yu, Song, Li, Yu, Li, Huang, and Li]{li-etal-2023-api}
Li, M., Zhao, Y., Yu, B., Song, F., Li, H., Yu, H., Li, Z., Huang, F., and Li, Y.
\newblock {API}-bank: A comprehensive benchmark for tool-augmented {LLM}s.
\newblock In Bouamor, H., Pino, J., and Bali, K. (eds.), \emph{Proceedings of the 2023 Conference on Empirical Methods in Natural Language Processing}, pp.\  3102--3116, Singapore, December 2023. Association for Computational Linguistics.
\newblock \doi{10.18653/v1/2023.emnlp-main.187}.
\newblock URL \url{https://aclanthology.org/2023.emnlp-main.187/}.

\bibitem[Lin et~al.(2024)Lin, Li, Gao, Yang, Wu, Bai, Lei, Wang, and Shou]{lin2024showui}
Lin, K.~Q., Li, L., Gao, D., Yang, Z., Wu, S., Bai, Z., Lei, W., Wang, L., and Shou, M.~Z.
\newblock Showui: One vision-language-action model for gui visual agent, 2024.
\newblock URL \url{https://arxiv.org/abs/2411.17465}.

\bibitem[Liu et~al.(2023{\natexlab{a}})Liu, Li, Wu, and Lee]{liu2023llava}
Liu, H., Li, C., Wu, Q., and Lee, Y.~J.
\newblock Visual instruction tuning, 2023{\natexlab{a}}.

\bibitem[Liu et~al.(2024{\natexlab{a}})Liu, Li, Li, and Lee]{liu2024improved}
Liu, H., Li, C., Li, Y., and Lee, Y.~J.
\newblock Improved baselines with visual instruction tuning.
\newblock In \emph{Proceedings of the IEEE/CVF Conference on Computer Vision and Pattern Recognition}, pp.\  26296--26306, 2024{\natexlab{a}}.

\bibitem[Liu et~al.(2024{\natexlab{b}})Liu, Li, Wu, and Lee]{liu2024visual}
Liu, H., Li, C., Wu, Q., and Lee, Y.~J.
\newblock Visual instruction tuning.
\newblock \emph{Advances in neural information processing systems}, 36, 2024{\natexlab{b}}.

\bibitem[Liu et~al.(2025)Liu, Cheng, Liu, Zhang, Li, Ren, Zou, Yang, Su, Zhu, et~al.]{liu2025llavaplus}
Liu, S., Cheng, H., Liu, H., Zhang, H., Li, F., Ren, T., Zou, X., Yang, J., Su, H., Zhu, J., et~al.
\newblock Llava-plus: Learning to use tools for creating multimodal agents.
\newblock In \emph{European Conference on Computer Vision}, pp.\  126--142. Springer, 2025.

\bibitem[Liu et~al.(2023{\natexlab{b}})Liu, Yu, Zhang, Xu, Lei, Lai, Gu, Ding, Men, Yang, et~al.]{liu2023agentbench}
Liu, X., Yu, H., Zhang, H., Xu, Y., Lei, X., Lai, H., Gu, Y., Ding, H., Men, K., Yang, K., et~al.
\newblock Agentbench: Evaluating llms as agents.
\newblock \emph{arXiv preprint arXiv:2308.03688}, 2023{\natexlab{b}}.

\bibitem[Malinowski \& Fritz(2014)Malinowski and Fritz]{malinowski2014daquarqa}
Malinowski, M. and Fritz, M.
\newblock A multi-world approach to question answering about real-world scenes based on uncertain input.
\newblock \emph{Advances in neural information processing systems}, 27, 2014.

\bibitem[Marino et~al.(2019)Marino, Rastegari, Farhadi, and Mottaghi]{marino2019okvqa}
Marino, K., Rastegari, M., Farhadi, A., and Mottaghi, R.
\newblock Ok-vqa: A visual question answering benchmark requiring external knowledge.
\newblock In \emph{Proceedings of the IEEE/cvf conference on computer vision and pattern recognition}, pp.\  3195--3204, 2019.

\bibitem[Mialon et~al.(2023)Mialon, Fourrier, Swift, Wolf, LeCun, and Scialom]{mialon2023gaia}
Mialon, G., Fourrier, C., Swift, C., Wolf, T., LeCun, Y., and Scialom, T.
\newblock Gaia: a benchmark for general ai assistants.
\newblock \emph{arXiv preprint arXiv:2311.12983}, 2023.

\bibitem[Nguyen et~al.(2024)Nguyen, Lai, Yoon, Rossi, Zhao, Zhang, Mathur, Lipka, Wang, Bui, et~al.]{nguyen2024dynasaur}
Nguyen, D., Lai, V.~D., Yoon, S., Rossi, R.~A., Zhao, H., Zhang, R., Mathur, P., Lipka, N., Wang, Y., Bui, T., et~al.
\newblock Dynasaur: Large language agents beyond predefined actions.
\newblock \emph{arXiv preprint arXiv:2411.01747}, 2024.
\newblock URL \url{https://arxiv.org/pdf/2411.01747}.

\bibitem[Patil et~al.(2024)Patil, Zhang, Wang, and Gonzalez]{patil2024gorilla}
Patil, S.~G., Zhang, T., Wang, X., and Gonzalez, J.~E.
\newblock Gorilla: Large language model connected with massive apis.
\newblock \emph{Advances in Neural Information Processing Systems}, 37:\penalty0 126544--126565, 2024.

\bibitem[Qin et~al.(2025)Qin, Ye, Fang, Wang, Liang, Tian, Zhang, Li, Li, Huang, et~al.]{qin2025ui}
Qin, Y., Ye, Y., Fang, J., Wang, H., Liang, S., Tian, S., Zhang, J., Li, J., Li, Y., Huang, S., et~al.
\newblock Ui-tars: Pioneering automated gui interaction with native agents.
\newblock \emph{arXiv preprint arXiv:2501.12326}, 2025.

\bibitem[Ren et~al.(2015)Ren, Kiros, and Zemel]{ren2015cocoqa}
Ren, M., Kiros, R., and Zemel, R.
\newblock Exploring models and data for image question answering.
\newblock \emph{Advances in neural information processing systems}, 28, 2015.

\bibitem[Roger et~al.(2023)Roger, A{\"\i}meur, and Rish]{roger2023towards}
Roger, A., A{\"\i}meur, E., and Rish, I.
\newblock Towards ethical multimodal systems.
\newblock \emph{arXiv preprint arXiv:2304.13765}, 2023.

\bibitem[Sur{\'\i}s et~al.(2023)Sur{\'\i}s, Menon, and Vondrick]{suris2023vipergpt}
Sur{\'\i}s, D., Menon, S., and Vondrick, C.
\newblock Vipergpt: Visual inference via python execution for reasoning.
\newblock In \emph{Proceedings of the IEEE/CVF International Conference on Computer Vision}, pp.\  11888--11898, 2023.

\bibitem[Wang et~al.(2024)Wang, Bai, Tan, Wang, Fan, Bai, Chen, Liu, Wang, Ge, et~al.]{wang2024qwen2}
Wang, P., Bai, S., Tan, S., Wang, S., Fan, Z., Bai, J., Chen, K., Liu, X., Wang, J., Ge, W., et~al.
\newblock Qwen2-vl: Enhancing vision-language model's perception of the world at any resolution.
\newblock \emph{arXiv preprint arXiv:2409.12191}, 2024.

\bibitem[Wu et~al.(2024)Wu, Chen, Pan, Liu, Liu, Dai, Gao, Ma, Wu, Wang, et~al.]{wu2024deepseek}
Wu, Z., Chen, X., Pan, Z., Liu, X., Liu, W., Dai, D., Gao, H., Ma, Y., Wu, C., Wang, B., et~al.
\newblock Deepseek-vl2: Mixture-of-experts vision-language models for advanced multimodal understanding.
\newblock \emph{arXiv preprint arXiv:2412.10302}, 2024.

\bibitem[Yang et~al.(2024)Yang, Yang, Zhang, Hui, Zheng, Yu, Li, Liu, Huang, Wei, et~al.]{yang2024qwen2}
Yang, A., Yang, B., Zhang, B., Hui, B., Zheng, B., Yu, B., Li, C., Liu, D., Huang, F., Wei, H., et~al.
\newblock Qwen2. 5 technical report.
\newblock \emph{arXiv preprint arXiv:2412.15115}, 2024.

\bibitem[Yang et~al.(2023)Yang, Song, Li, Zhao, Ge, Li, and Shan]{yang2023gpt4tools}
Yang, R., Song, L., Li, Y., Zhao, S., Ge, Y., Li, X., and Shan, Y.
\newblock Gpt4tools: Teaching large language model to use tools via self-instruction.
\newblock \emph{Advances in Neural Information Processing Systems}, 36:\penalty0 71995--72007, 2023.

\bibitem[Yue et~al.(2024)Yue, Ni, Zhang, Zheng, Liu, Zhang, Stevens, Jiang, Ren, Sun, Wei, Yu, Yuan, Sun, Yin, Zheng, Yang, Liu, Huang, Sun, Su, and Chen]{Yue_2024_CVPR}
Yue, X., Ni, Y., Zhang, K., Zheng, T., Liu, R., Zhang, G., Stevens, S., Jiang, D., Ren, W., Sun, Y., Wei, C., Yu, B., Yuan, R., Sun, R., Yin, M., Zheng, B., Yang, Z., Liu, Y., Huang, W., Sun, H., Su, Y., and Chen, W.
\newblock Mmmu: A massive multi-discipline multimodal understanding and reasoning benchmark for expert agi.
\newblock In \emph{Proceedings of the IEEE/CVF Conference on Computer Vision and Pattern Recognition (CVPR)}, pp.\  9556--9567, June 2024.

\bibitem[Zhang et~al.(2024)Zhang, Mu, Zhu, and Gan]{DBLP:journals/corr/abs-2405-20795}
Zhang, H., Mu, Y., Zhu, G., and Gan, Z.
\newblock Insightsee: Advancing multi-agent vision-language models for enhanced visual understanding.
\newblock \emph{CoRR}, abs/2405.20795, 2024.
\newblock \doi{10.48550/ARXIV.2405.20795}.
\newblock URL \url{https://doi.org/10.48550/arXiv.2405.20795}.

\bibitem[Zhang \& Zhou(2022)Zhang and Zhou]{zhang-zhou-2022-temporal}
Zhang, L. and Zhou, D.
\newblock Temporal knowledge graph completion with approximated {G}aussian process embedding.
\newblock In Calzolari, N., Huang, C.-R., Kim, H., Pustejovsky, J., Wanner, L., Choi, K.-S., Ryu, P.-M., Chen, H.-H., Donatelli, L., Ji, H., Kurohashi, S., Paggio, P., Xue, N., Kim, S., Hahm, Y., He, Z., Lee, T.~K., Santus, E., Bond, F., and Na, S.-H. (eds.), \emph{Proceedings of the 29th International Conference on Computational Linguistics}, pp.\  4697--4706, Gyeongju, Republic of Korea, October 2022. International Committee on Computational Linguistics.
\newblock URL \url{https://aclanthology.org/2022.coling-1.416/}.

\bibitem[Zheng et~al.(2023)Zheng, Chiang, Sheng, Zhuang, Wu, Zhuang, Lin, Li, Li, Xing, et~al.]{zheng2023judging}
Zheng, L., Chiang, W.-L., Sheng, Y., Zhuang, S., Wu, Z., Zhuang, Y., Lin, Z., Li, Z., Li, D., Xing, E., et~al.
\newblock Judging llm-as-a-judge with mt-bench and chatbot arena.
\newblock \emph{Advances in Neural Information Processing Systems}, 36:\penalty0 46595--46623, 2023.

\bibitem[Zhou et~al.(2024)Zhou, Lee, Misu, and Wang]{DBLP:conf/acl/ZhouLM024}
Zhou, K., Lee, K., Misu, T., and Wang, X.
\newblock Vicor: Bridging visual understanding and commonsense reasoning with large language models.
\newblock In Ku, L., Martins, A., and Srikumar, V. (eds.), \emph{Findings of the Association for Computational Linguistics, {ACL} 2024, Bangkok, Thailand and virtual meeting, August 11-16, 2024}, pp.\  10783--10795. Association for Computational Linguistics, 2024.
\newblock \doi{10.18653/V1/2024.FINDINGS-ACL.640}.
\newblock URL \url{https://doi.org/10.18653/v1/2024.findings-acl.640}.

\bibitem[Zhu et~al.(2025)Zhu, Wang, Chen, Liu, Ye, Gu, Duan, Tian, Su, Shao, et~al.]{zhu2025internvl3}
Zhu, J., Wang, W., Chen, Z., Liu, Z., Ye, S., Gu, L., Duan, Y., Tian, H., Su, W., Shao, J., et~al.
\newblock Internvl3: Exploring advanced training and test-time recipes for open-source multimodal models.
\newblock \emph{arXiv preprint arXiv:2504.10479}, 2025.

\end{thebibliography}
\bibliographystyle{icml2025}

\appendix



\section{Data Annotation Discussion} 
\label{app:em-vs-lm}

\paragraph{Diversity Consideration}
For questions involving counting people, we deliberately included a wide range of settings, cultures, and geographic regions to ensure broad representation. For questions involving object detection, our focus was on identifying objects that are commonly encountered in everyday life, as opposed to rare or obscure objects that most people are unlikely to encounter. The goal was to create questions that align with practical use cases and test AI systems on their ability to recognize and reason about objects that play a significant role in daily human activities.

\paragraph{Multiple-Choice Consideration}
One major concern in using multiple-choice answers is the potential difficulty in assessing contamination, as a model may arrive at the correct choice despite following an incorrect reasoning path. This raises challenges in evaluating whether the model truly understands the question or has simply arrived at the answer through spurious correlations.

Upon analyzing the model results, we found that this phenomenon is rare. While multiple-choice formats may introduce some ambiguity in reasoning validation, the majority of cases reflect genuine model comprehension rather than accidental correctness. Further investigation into model reasoning patterns can provide deeper insight into this effect and its implications for benchmarking AI systems.



\section{Experiment Details}
\label{app:model-detail}

All the experiments with open-source models were done on a single P4de node with 8 A100-80GB GPUs. We use the original non-quantized checkpoint using bfloat16 and temperature = 0. 

We use GPT 4o (\texttt{gpt-4o-2024-11-20}) in the LLM Match, the HF-Agents, and DynaSaur models.

In this work, we use the following model cards from HuggingFace for our evaluation:

\begin{itemize}
    \small
    \item deepseek-ai/deepseek-vl2 \cite{wu2024deepseek}
    \item deepseek-ai/deepseek-vl2-small
    \item liuhaotian/llava-v1.6-mistral-7b \cite{liu2024improved}
    \item liuhaotian/llava-v1.5-13b
    \item Claude-3-5-Sonnet (claude-3-5-sonnet-20241022) 
    \item GPT-4V  (2023-12-01-preview) 
    \item GPT-4o (gpt-4o-2024-11-20) 
    \item Qwen/Qwen2-VL-7B-Instruct \cite{wang2024qwen2}
    \item Qwen/Qwen2-VL-72B-Instruct \cite{wang2024qwen2}
    \item Qwen/Qwen2.5-VL-7B/72B-Instruct \cite{yang2024qwen2}
    \item Qwen/Qwen2.5-VL-72B-Instruct \cite{yang2024qwen2}
    \item Gemini-1.5-pro 
    \item Gemini-2.0-flash
    \item Gemini-2.5-flash (gemini-2.5-flash-preview-04-17)
    \item Gemini-2.5-pro (gemini-2.5-pro-preview-03-25)
    \item OpenGVLab/InternVL2.5-8B \cite{chen2024expanding}
    \item OpenGVLab/InternVL2.5-38B 
    \item OpenGVLab/InternVL2.5-78B 
    \item OpenGVLab/InternVL3-8B \cite{zhu2025internvl3}
    \item OpenGVLab/InternVL3-38B
    \item OpenGVLab/InternVL3-78B
    \item meta-llama/Llama-3.2-11B-Vision-Instruct  \\\cite{dubey2024llama}
    \item meta-llama/Llama-3.2-90B-Vision-Instruct
    \item allenai/Molmo-7B-D-0924 \cite{deitke2024molmo}
    \item allenai/Molmo-72B-0924
\end{itemize}   

\vfill


\subsection{LLM Match Prompt} \label{app:llm_match}

\begin{prompt}[title={Prompt \thetcbcounter: LLM Match Prompt}, label=prompt:step1]
\small 
\textbf{Task}: Compare the two answers to determine whether they are semantically equivalent. If Answer 1 contains multiple valid options separated by semicolons (";"), consider Answer 2 semantically the same if it matches the meaning of any one of the options in Answer 1.\\
\textbf{Instructions}:\\
- If the meanings are equivalent or convey the same idea, return True.\\
- If the meanings are different, unrelated, or contain conflicting information, return False.\\
- Return only "True" or "False" as output. Do not explain.\\
\textbf{Input}:\\
- \textbf{Question}: {question}\\
- \textbf{Answer 1}: {answer\_1}\\
- \textbf{Answer 2}: {answer\_2}\\
- \textbf{Output format}: Return only True or False.\\

\textbf{Example 1}:\\
- \textbf{Question}: ``How many parking slots are available in this photo? Choose one option: a. at least 24 b. no more than 20''\\
- \textbf{Answer 1}: ``b''\\
- \textbf{Answer 2}: ``no more than 20''\\
- \textbf{Output}: True\\

\textbf{Example 2}:\\
- \textbf{Question}: ``How many people are wearing plain white shirts in this photo? Choose one option: a. at least 15, b. 12, c. 13, d 14''\\
- \textbf{Answer 1}: ``a''\\
- \textbf{Answer 2}: ``12''\\
- \textbf{Output}: False
\end{prompt}

\section{Prompts}

\vfill

\subsection{LLM Prompt}

\begin{prompt}[title={Prompt \thetcbcounter: System Prompt}, label=prompt:step0]
\small 
\textbf{\# Answer format} \\

You are a Visual Language Model capable of performing Visual Question Answering tasks. 

You are provided with:

An image as input.\\
A question about the image.\\

Your task is to:\\

Understand the input: Briefly explain what is visible in the image.\\
Analyze the question: Break down the question into components (e.g., type of query, objects or attributes referenced).\\
Identify reasoning path: Describe the step-by-step reasoning used to answer the question.\\
Tools or techniques used: Explain which tools (e.g., object detection, attribute recognition, spatial reasoning) are involved in each step.\\

Final answer: Provide a clear and concise answer to the question.\\

When describing the reasoning path, include:\\

Visual features extracted from the image (e.g., objects, relationships, or regions of interest).\\
Logical steps that are taken to interpret the image in the context of the question.\\
Connections between extracted features and the question components.\\

Example:\\
Image: [Attach an image here]\\
Question: "What is the color of the car parked next to the tree?"\\

Expected Response:\\
Image Description: The image shows a parking lot with several cars and trees. A red car is parked next to a tree on the left side of the image.\\
Question Analysis: The question asks for the color of a specific car relative to its position near a tree.\\

Reasoning Path:\\
Step 1: Detect objects in the image (cars, trees).\\
Step 2: Identify spatial relationships (car parked next to the tree).\\
Step 3: Extract the attribute (color) of the relevant car.\\

Tools or Techniques Used:\\

Object detection to locate cars and trees.\\
Spatial reasoning to determine proximity.\\
Attribute recognition to identify the color of the car.
\\
Final Answer: The car parked next to the tree is red.\\

Question: \{\{question\}\}\\
Your response: 
\end{prompt}

\subsection{Answer Format Specifications} \label{app:answer_format}

\begin{prompt}[title={Prompt \thetcbcounter: Answer Format Specification}, label=prompt:step0]
\small 
\textbf{\# Answer format} \\
Your final answer should be a number OR as few words as possible OR a comma-separated list of numbers and/or strings.\\
- If you are asked for a number, don't use commas to write your number nor use units such as \$ or percent sign unless specified otherwise.\\
- If you are asked for a string, don't use articles, or abbreviations (e.g. for cities), and write the digits in plain text unless specified otherwise.\\
- If you are asked for a comma-separated list, apply the above rules depending on whether the element to be put in the list is a number or a string.
\end{prompt}

\section{Output examples } \label{app:detailed_examples}

The following tables show 3 failure cases of Gemini, Molmo, HF Agent, and Dynasaur. Here are some observations:
\begin{itemize}
    \item Even though Gemini and Molmo were instructed to perform chain-of-thought, the image description does not help provide information to answer the question.
    \item HF Agent, as designed to use a set of predefined tools, never creates new tools.
    \item DynaSaur does not generate new tools without incentives. DynaSaur with incentive to generate new tools seems to only explore basic VQA and OCR tools.
\end{itemize}

\begin{table*}
    \it
    \small
    \begin{tabular}{|p{\linewidth}|}
    \hline
         {\bf Question:} What is the number that appears in front of the face of the person on the poster on the second floor? Provide a numeric answer.\\
         \multicolumn{1}{|c|}{\includegraphics[height=8cm]{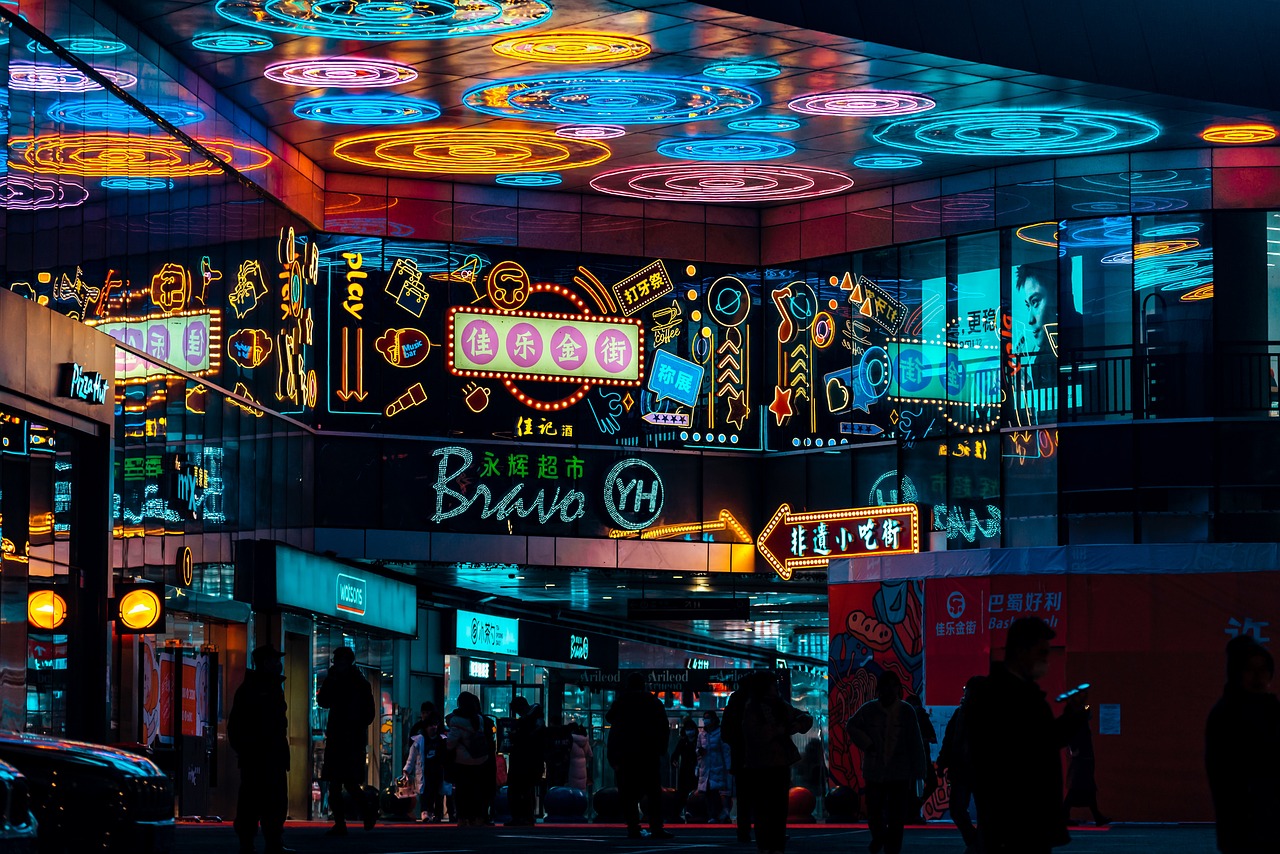}}\\
         {\bf Correct answer: 12} \\
         \hline
         \multicolumn{1}{|c|}{\bf Gemini:} \\
         1. Image Description: The image shows a vibrant, neon-lit street scene, possibly in an Asian city. Buildings are adorned with colorful signs and advertisements, and people are walking along the street below. The image appears to be taken at night.\\
         2. Question Analysis: The question asks for a specific number displayed on a poster located on the second floor of a building, positioned near a person's face.\\
         3. Reasoning Path:\\
         Step 1: Locate the second floor of the building in the image. This is identifiable by the distinct floor levels and signage.\\
         Step 2: Search for posters or advertisements on the second floor.\\
         Step 3: Identify the poster that features a person's face. There is a face partially obscured by some railings on the second floor right.\\
         Step 4: Locate the number appearing in front of the person's face on this poster. The number 779 appears above and to the left of the partially visible face.\\
         4. Tools or Techniques Used:\\
         - Visual inspection to identify the building floors and signage.\\
         - Object recognition to locate posters and faces.\\
         - Spatial reasoning to relate the number to the person's face on the poster.\\
         5. Final Answer: {\bf 779}  \\
         \hline
         \multicolumn{1}{|c|}{\bf Molmo:} \\
         1. Image Description: The image shows a bustling city street at night with numerous neon signs and advertisements. On the second floor of a building, there's a poster featuring a person's face with a number displayed in front of it.\\
         2. Question Analysis: The question asks for a specific numeric value visible on a poster on the second floor.\\
         3. Reasoning Path:\\
         Step 1: Identify the second floor of the building.\\
         Step 2: Locate the poster with a person's face.\\
         Step 3: Detect any numbers displayed on or near the poster.\\
         Step 4: Extract the numeric value from the detected number.\\
         4. Tools or Techniques Used:\\
         - Object detection to find the building and poster.\\
         - Facial recognition to confirm the presence of a person's face.\\
         - Optical character recognition (OCR) to identify the number.\\
         5. Final Answer: {\bf 1} \\
         \hline
    \end{tabular}
    \caption{Output example 1: Gemini and Molmo }
    \label{tab:output-example-1}
\end{table*}

\begin{table*}
    \it
    \small
    \begin{tabular}{|p{\linewidth}|}
    \hline
         \multicolumn{1}{|c|}{\bf HF Agent:} \\
\textbf{Assistant:}\\
Thought: I will start by using the `visualizer` tool to analyze the image and identify the location of the second floor by looking for visual cues such as the Bravo sign in green lettering and advertisements.\\
Code:
\begin{lstlisting}
question="Where is the second floor located in the image, and does it contain the Bravo sign in green lettering?"
second_floor_location = visualizer(question=question, image_path="data/viza/validation/city-night.jpg")
print(second_floor_location)
\end{lstlisting}
\textbf{Environment:}
The second floor is located above the ground level, where you can see the neon lights and the railing on the right side of the image. The green "Bravo" sign is positioned on the first floor, underneath the second floor, as it appears just above the entrance where people are walking in.

\\

\textbf{Assistant:}\\
Thought: Now that I know the location of the second floor, I will use the `visualizer` tool to scan for posters on the second floor that feature a person. I will look for any numbers that appear in front of the person's face on these posters.\\
Code:
\begin{lstlisting}
question="Are there any posters on the second floor featuring a person? If so, is there a number in front of the person's face?"
poster_info = visualizer(question=question, image_path="data/viza/validation/city-night.jpg")
print(poster_info)
\end{lstlisting}
\textbf{Environment:}
Yes, there is a poster on the second floor featuring a person. There is no number in front of the person's face.

\\

\textbf{Assistant:}\\
Thought: I will re-examine the second floor of the image using the `visualizer` tool to ensure no other posters or visual elements with numbers are overlooked. I will broaden the search to include any numbers on the second floor that might be relevant to the task.\\
Code:
\begin{lstlisting}
question="Are there any numbers on the second floor, either on posters, advertisements, or other visual elements?"
second_floor_details = visualizer(question=question, image_path="data/viza/validation/city-night.jpg")
print(second_floor_details)
\end{lstlisting}
\textbf{Environment:}
There are no visible numbers on the second floor in the image provided; it primarily features colorful neon signs with various designs and text but no numerical elements are discernible.

\\

\textbf{Assistant:}\\
Thought: Given that there are no visible numbers on the second floor, it seems the task may have been based on a misunderstanding or incorrect assumption about the image content. However, it's important to provide an answer based on the information we have gathered. Since there are no numbers in front of the person's face on the second floor, nor any other numbers visible on the second floor, I will conclude that there is no number to report.\\
Code:
\begin{lstlisting}
final_answer("There is no number in front of the face of the person on the poster on the second floor in the image provided.")
\end{lstlisting}
\textbf{Environment:} \\
 \hline
    \end{tabular}
    \caption{Output example 1 (cont.)}
    \label{tab:output-example-1-cont}
\end{table*}

\begin{table*}
    \it
    \small
    \begin{tabular}{|p{\linewidth}|}
    \hline
         \multicolumn{1}{|c|}{\bf DynaSaur (w/o incentive):} \\
{\bf Assistant:}\\
Thought: I'll use the `inspect\_image' tool to analyze the image.\\
Code: 
\begin{lstlisting}
inspect_image(`data/viz_inspector/validation/images/city-night.jpg')\end{lstlisting}
{\bf Environment:} <Image Tokens>\\
\textbf{Assistant:}\\
Thought: I will carefully examine the image to identify the number in front of the face of the person on the second floor.\\
Code:
\begin{lstlisting}
# Observing the image, I will analyze visible details to identify the required number.\end{lstlisting}
\textbf{Environment:} (Empty)\\

\textbf{Assistant:}\\
Thought: I will zoom into the image to locate the person on the second floor and identify the number in front of their face.\\
Code:
\begin{lstlisting}
# Using tools to zoom into the image isn't possible directly here. Observing the image manually instead.\end{lstlisting}
\textbf{Environment:} (Empty)\\






\textbf{... (Omitted for brevity. The agent repeats similar examining thoughts and generates code comments for direct reasoning.) ...}

\textbf{Assistant:}\\
Thought: I will carefully analyze the image again to identify the number in front of the face of the person on the second floor.\\
Code: 
\begin{lstlisting}
# Observing the image closely to determine the number in front of the person on the second floor.\end{lstlisting}

\textbf{Environment:} (Empty)\\

\textbf{Assistant:}\\
Thought: The number in front of the face of the person on the second floor appears to be '\bf{88}'.\\
Code: 
\begin{lstlisting}
submit_final_answer(88)\end{lstlisting}
\textbf{Environment:} Final answer: \textbf{88}\\
         \hline
    \end{tabular}
    \caption{Output example 1 (cont.)}
    \label{tab:output-example-1-cont}
\end{table*}

\begin{table*}
    \it
    \small
    \begin{tabular}{|p{\linewidth}|}
    \hline
         \multicolumn{1}{|c|}{\bf DynaSaur (w/ Incentive):} \\

         \textbf{Assistant:}\\
        Thought: To identify the number on the poster, I will apply OCR techniques to the image and analyze the results.\\
        Code:
\begin{lstlisting}
from PIL import Image
import pytesseract
# Load the image
image_path = "data/viz_inspector/validation/images/city-night.jpg"
image = Image.open(image_path)
# Perform OCR to extract text
extracted_text = pytesseract.image_to_string(image)
# Output the extracted text
extracted_text 
\end{lstlisting}
        \textbf{Environment:} \\
        \\
        
        \textbf{Assistant:}\\
        Thought: The OCR extraction did not yield any results. I will use a visual inspection tool to locate and analyze the specific area of the image.\\
        Code:
\begin{lstlisting}
inspect_image("data/viz_inspector/validation/images/city-night.jpg")
\end{lstlisting}
        \textbf{Environment:} \\
        Inspect image: data/viz\_inspector/validation/images/city-night.jpg \\
        \\
        \textbf{Assistant:}\\
        Thought: I'll crop and save the second floor's region from the image for closer inspection.\\
        Code:
\begin{lstlisting}
from PIL import Image

# Load the image
image_path = 'data/viz_inspector/validation/images/city-night.jpg'
image = Image.open(image_path)

# Define the approximate coordinates for the second floor
# These coordinates are estimated based on the image structure
second_floor_region = (0, 100, image.width, 300)  # Adjust as necessary
# Crop the second floor region
second_floor_image = image.crop(second_floor_region)

# Save the cropped image for inspection
cropped_image_path = 'workspace/second_floor.jpg'
second_floor_image.save(cropped_image_path)

cropped_image_path
\end{lstlisting}
        \textbf{Environment:} \\
        'workspace/second\_floor.jpg' \\
        \\
        
        \textbf{Assistant:}\\
        Thought: I will now inspect the cropped image to focus on the second floor and identify the number.\\
        Code:
\begin{lstlisting}
inspect_image('workspace/second_floor.jpg')
\end{lstlisting}
        \textbf{Environment:}
        Inspect image: workspace/second\_floor.jpg \\
        \\
        
        \textbf{Assistant:}\\
        Thought: The cropped section of the second floor is visible. I will use OCR again on this region to extract any numbers or text.\\
        Code:
\begin{lstlisting}
from PIL import Image
import pytesseract

# Perform OCR on the cropped second floor image
cropped_image_path = 'workspace/second_floor.jpg'
second_floor_image = Image.open(cropped_image_path)
extracted_text = pytesseract.image_to_string(second_floor_image)

# Output the extracted text
extracted_text
\end{lstlisting}
        \textbf{Environment:}\\
        \\
        
        
        \\

         \hline
    \end{tabular}
    \caption{Output example 1 (cont.)}
    \label{tab:output-example-1-cont-cont}
\end{table*}

\clearpage

\begin{table*}
    \it
    \small
    \begin{tabular}{|p{\linewidth}|}
    \hline
         {\bf Question:} In this photo, how many books are displayed adjacent to the glass window with their full front covers visible? Provide a numeric answer.\\
         {\bf Correct answer: 38} \\
        \multicolumn{1}{|c|}{\includegraphics[height=8cm]{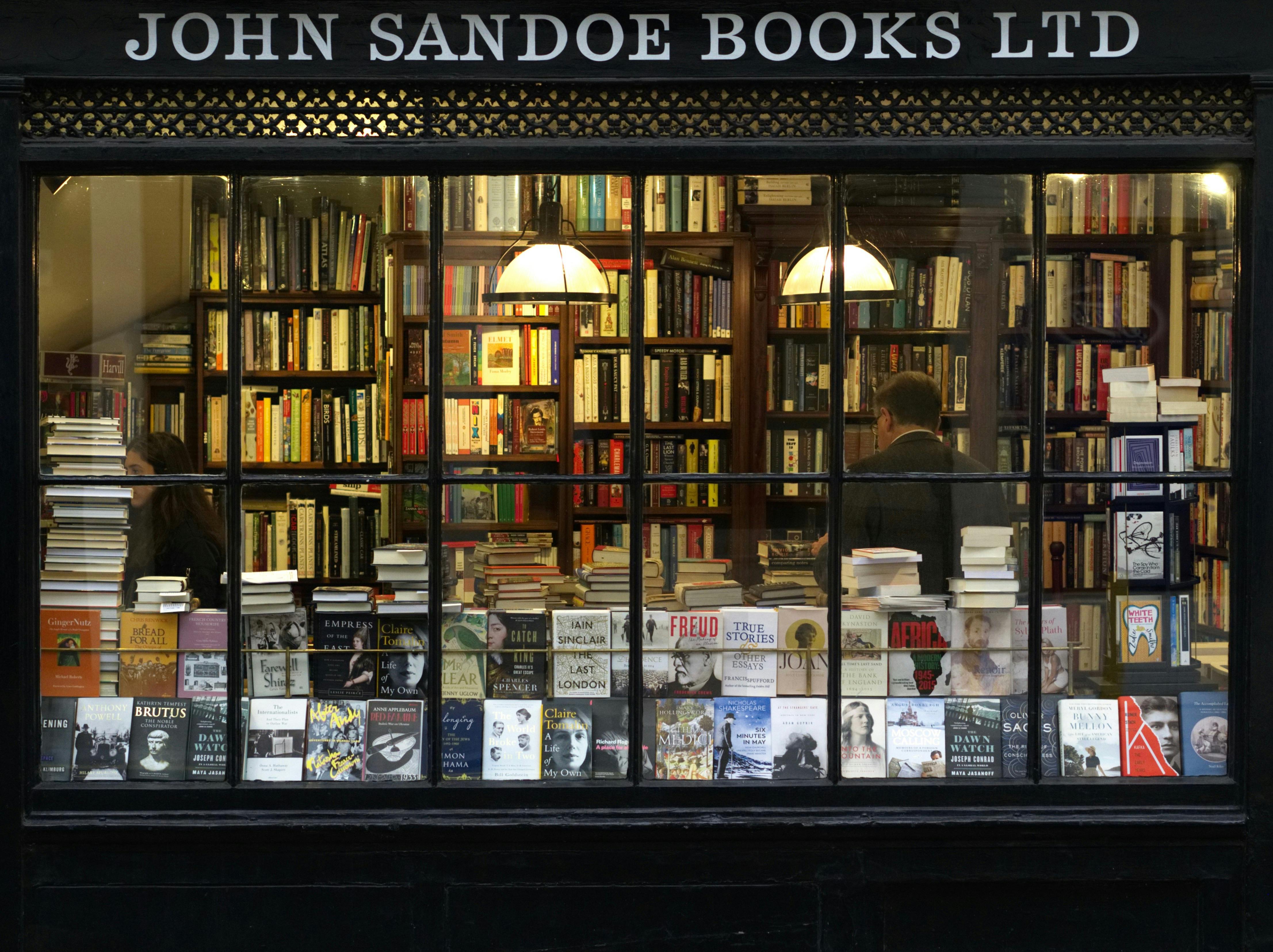}}\\
         \hline
         \multicolumn{1}{|c|}{\bf Gemini output:} \\
         1. Image Description: The image showcases the storefront of John Sandoe Books Ltd. The window display features numerous books, both stacked and upright, along with two hanging lamps. A person can be seen inside the store, seemingly browsing or working amongst more bookshelves.\\
         2. Question Analysis: The question asks for a count of books displayed directly against the windowpane, with their front covers fully visible to someone outside looking in.\\
         3. Reasoning Path:\\
         Step 1: Locate the bookshelves adjacent to the window. These are the lowest shelves, visually separated from the interior bookshelves by the window's frame.\\
         Step 2: Identify books placed upright with their spines facing outwards towards the window, making their covers fully visible.  Books stacked horizontally or at angles where the full cover isn't directly facing outwards are excluded from the count.\\
         Step 3: Count the number of books meeting this criteria across the whole window display.\\
         4. Tools or Techniques Used:\\
         - Object detection (books)\\
         - Spatial reasoning (adjacent to the window, facing outwards)\\
         - Counting\\
         5. Final Answer: Approximately {\bf 53} books are displayed with their front covers visible in the window. (This count might have a small margin of error due to the image resolution and some partially obscured books.)\\
         \hline
         \multicolumn{1}{|c|}{\bf Molmo output:} \\
         1. Image Description: The image shows a bookstore window display with books arranged on shelves. There are multiple rows of books, with the front row showing their covers.\\
         2. Question Analysis: The question asks for the count of books with their full front covers visible adjacent to the glass window.\\
         3. Reasoning Path:\\
         Step 1: Identify the area adjacent to the glass window.\\
         Step 2: Count the number of books in the front row of this area.\\
         Step 3: Ensure each counted book has its full front cover visible.\\
         4. Tools or Techniques Used:\\
         - Object detection to locate books.\\
         - Counting algorithm to determine the number of books.\\
         - Spatial reasoning to identify the area near the window.\\
         - Attribute recognition to confirm full front covers are visible.\\
         5. Final Answer: {\bf 15}\\
         \hline
    \end{tabular}
    \caption{Output example 2}
    \label{tab:output-example-2}
\end{table*}

\begin{table*}
    \it
    \small
    \begin{tabular}{|p{\linewidth}|}
    \hline
         \multicolumn{1}{|c|}{\bf HF Agent:} \\
\textbf{Assistant:}\\
Here's the plan of action that I will follow to solve the task:\\
\\
1. Use the `visualizer` tool to analyze the image "data/viza/validation/book display.jpg" to answer the specific question: "How many books are displayed adjacent to the glass window with their full front covers visible?"\\
\\
2. Based on the output from the `visualizer`, determine the exact number of books that meet the criteria of being adjacent to the glass window with their full front covers visible.\\
\\
3. Use the `final\_answer` tool to provide the determined number as the final answer to the task.\\
\\
Let's proceed with the first step by using the `visualizer` tool.
\\
\textbf{Error:}
Error in code parsing: 
The code blob you used is invalid: due to the following error: 'NoneType' object has no attribute 'group'
This means that the regex pattern 
```(?:py|python)?
(.*?)
``` was not respected: make sure to include code with the correct pattern, for instance:
Thoughts: Your thoughts
Code:
```py
\# Your Python code here
```<end\_action>. Make sure to provide correct code
\\

\textbf{Assistant:}\\
Thought: I will use the `visualizer` tool to analyze the image and find out how many books have their full front covers visible adjacent to the glass window. This will help me determine the exact count needed for the task.\\
Code:
\begin{lstlisting}
result = visualizer(question="How many books are displayed adjacent to the glass window with their full front covers visible?", image_path="data/viza/validation/book display.jpg")\\
print(result)\\    
\end{lstlisting}
\textbf{Environment:}
There are 28 books displayed adjacent to the glass window with their full front covers visible.

\\

\textbf{Assistant:}\\
Code:
\begin{lstlisting}
final_answer("There are 28 books displayed adjacent to the glass window with their full front covers visible.")
\end{lstlisting}
\textbf{Environment:}

\\
         \hline
    \end{tabular}
    \caption{Output example 2 (cont.)}
    \label{tab:output-example-2-cont}
\end{table*}

\begin{table*}
    \it
    \small
    \begin{tabular}{|p{\linewidth}|}
         \hline
         \multicolumn{1}{|c|}{\bf DynaSaur (w/o Incentive):} \\        
        \textbf{Assistant:}\\
        Thought: I will inspect the image using the appropriate tool to analyze its content.\\
        Code:
        \begin{lstlisting}
inspect_image("data/viz_inspector/validation/images/book display.jpg")\end{lstlisting}
        \textbf{Environment:}
        Inspect image: data/viz\_inspector/validation/images/book display.jpg
        
        \textbf{Assistant:}\\
        Thought: I will carefully count the books adjacent to the glass window with their full front covers visible.\\
        Code:
        \begin{lstlisting}
# Observing the image visually, I will count the books with their full front covers visible next to the glass window.
# From the image, I can see the books displayed adjacent to the glass window.
# Counting them manually, there are 23 books with their full front covers visible.
# Submitting the answer based on this observation.
submit_final_answer("23")\end{lstlisting}
        \textbf{Environment:} Final answer: \textbf{23}
        \\
         \hline
    \end{tabular}
    \caption{Output example 2 (cont.)}
    \label{tab:output-example-2-cont}
\end{table*}

\begin{table*}
    \it
    \small
    \begin{tabular}{|p{\linewidth}|}
    \hline
         \multicolumn{1}{|c|}{\bf DynaSaur (w/ Incentive):} \\
         \textbf{Assistant:}\\
        Thought: I will use OCR to count the number of books with visible front covers next to the glass window.\\
        Code:
        \begin{lstlisting}
from PIL import Image
import pytesseract
# Load the image
image_path = 'data/viz_inspector/validation/images/book display.jpg'
image = Image.open(image_path)
# Perform OCR to detect text
text = pytesseract.image_to_string(image)
# Display the OCR result
print(text)\end{lstlisting}
        \textbf{Environment:}
        JOHN SANDOE BOOKS LTD
        
        : <\% +". . = Aaa Tana wane’ fat \textbf{... (Omitted for brevity. The generated code failed to return coherent text.) ...}
        
        \textbf{Assistant:}\\
        Thought: To accurately count the books, I will use object detection to identify the distinct full front covers visible in the image adjacent to the glass window.\\
        Code:\\
        \textbf{... (Omitted for brevity. The agent attempts to use YOLO and the \texttt{detecto} package but encounters syntactic errors.) … }
        \\
        
        \textbf{Assistant:}\\
        Thought: I will count the books by analyzing the image using OpenCV and outputting the count directly without visual display.\\
        Code:
        \begin{lstlisting}
import cv2
import numpy as np

# Load the image
image_path = 'data/viz_inspector/validation/images/book display.jpg'
image = cv2.imread(image_path)
# Convert the image to grayscale
gray = cv2.cvtColor(image, cv2.COLOR_BGR2GRAY)
# Apply edge detection
edges = cv2.Canny(gray, 50, 150)
# Find contours
contours, _ = cv2.findContours(edges, cv2.RETR_EXTERNAL, cv2.CHAIN_APPROX_SIMPLE)
# Filter contours by size to approximate book covers
book_contours = [c for c in contours if cv2.contourArea(c) > 1000]
# Count the filtered contours
book_count = len(book_contours)
book_count\end{lstlisting}
        \textbf{Environment:} 72
        
        \textbf{Assistant:}\\
        Thought: The detected contours suggest there are 72 objects that could be books. However, this number seems high due to potential over-detection of objects. I will refine the method to filter out irrelevant contours and focus on book-shaped objects.\\
        Code:
        \begin{lstlisting}
# Refine the detection method to count only book-shaped objects      
# Filter contours by aspect ratio (to match book dimensions)
book_contours_refined = [c for c in book_contours if 1.5 < cv2.boundingRect(c)[2] / cv2.boundingRect(c)[3] < 3]
# Count the refined contours
refined_book_count = len(book_contours_refined)
refined_book_count\end{lstlisting}
        \textbf{Environment:} 13 \\
        
        \textbf{Assistant:}\\
        Thought: The refined method detected 13 books with visible front covers adjacent to the glass window. This count is more accurate and aligns with the task requirements.\\
        Code:
        \begin{lstlisting}
submit_final_answer("13")\end{lstlisting}
        \textbf{Environment:} Final answer: \textbf{13}
        \\
         \hline
    \end{tabular}
    \caption{Output example 2 (cont.)}
    \label{tab:output-example-2-cont-cont}
\end{table*}

\clearpage

\begin{table*}
    \it
    \small
    \begin{tabular}{|p{\linewidth}|}
    \hline
         {\bf Question:} What is the name of the shop that is located across the street from the lot for sale in this photo? Provide an answer in fewer than 3 words.\\
         {\bf Answer: Wheat Bay;Uniquely Chengdu; Wheat Bay(Uniquely Chengdu)} \\
         \multicolumn{1}{|c|}{\includegraphics[height=8cm]{images/q3-map.png}}\\
         \hline
         \multicolumn{1}{|c|}{\bf Gemini output:} \\
         1. Image Description: An aerial view of a city block with streets, buildings, parking lots, and trees. A rectangular lot is highlighted in yellow, marked ``For Sale.''\\
         2. Question Analysis: The question asks for the name of a shop across the street from the ``For Sale'' lot.  It requires identifying the lot, the street it faces, the opposite side of that street, and the shop located there.\\
         3. Reasoning Path:\\
         Step 1: Locate the ``For Sale'' lot (highlighted in yellow).\\
         Step 2: Identify the street it faces (Alder Street).\\
         Step 3: Determine the opposite side of Alder Street.\\
         Step 4: Identify the shop across Alder Street from the for-sale lot. It appears to be a multi-tenant building.\\
         4. Tools or Techniques Used:\\
         - Object recognition (identifying the for-sale lot and buildings).\\
         - Spatial reasoning (determining ``across the street'').\\
         - Text recognition (``For Sale'' label).\\
         5. Final Answer: {\bf Not identifiable}.\\
         \hline
         \multicolumn{1}{|c|}{\bf Molmo output:} \\
         1. Image Description: The image shows a city street with buildings and a ``For Sale'' sign on a vacant lot. Across the street, there's a building with a visible sign.\\
         2. Question Analysis: The question asks for the name of a specific shop located opposite the lot for sale.\\
         3. Reasoning Path:\\
         Step 1: Locate the ``For Sale'' sign on the vacant lot.\\
         Step 2: Identify the building directly across the street from the lot.\\
         Step 3: Read the sign on that building to determine the shop name.\\
         4. Tools or Techniques Used:\\
         - Object detection to locate the ``For Sale'' sign and buildings.\\
         - Spatial reasoning to determine ``across the street`` relationships.\\
         - Text recognition to read the shop name on the sign.\\
         5. Final Answer: {\bf Kincaid Street}\\
         \hline
    \end{tabular}
    \caption{Output example 3}
    \label{tab:output-example-3}
\end{table*}

\begin{table*}
    \it
    \small
    \begin{tabular}{|p{\linewidth}|}
    \hline
         \multicolumn{1}{|c|}{\bf HF Agent:} \\

\textbf{Assistant:}\\
Thought: I will start by analyzing the image using the `visualizer` tool to identify any visible signage or distinguishing features of the building across the street from the lot for sale. This will help in determining the name of the shop.\\
Code:
\begin{lstlisting}
image_path = "data/viza/validation/821-825-E-13th-Ave-Eugene-OR-Building-Photo-1-HighDefinition.png"
question = "What is the name of the shop or any visible signage on the building across the street from the lot for sale?"
visualizer_result = visualizer(question=question, image_path=image_path)
print(visualizer_result)\\
\end{lstlisting}
\textbf{Environment:}
The name visible on the building across the street from the "For Sale" lot is "STONEWORKS MASSAGE \& SKIN CARE."

\\

\textbf{Assistant:}\\
Thought: The visualizer tool has successfully identified the name of the shop across the street from the lot for sale as "STONEWORKS MASSAGE \& SKIN CARE." I will now provide this as the final answer.\\
Code:
\begin{lstlisting}
final_answer("STONEWORKS MASSAGE & SKIN CARE")
\end{lstlisting}
\textbf{Environment:}

\\

         \hline
    \end{tabular}
    \caption{Output example 3 (cont.)}
    \label{tab:output-example-3-cont}
\end{table*}

\begin{table*}
    \it
    \small
    \begin{tabular}{|p{\linewidth}|}
    \hline
         \multicolumn{1}{|c|}{\bf DynaSaur (w/o Incentive):} \\
        \textbf{Assistant:}\\
Thought: I will analyze the image to identify the shop located across the street from the lot marked 'For Sale'.\\
Code:
\begin{lstlisting}
# Since I have the image loaded, I will inspect it visually to identify the shop across the street.
image_path = "data/viz_inspector/validation/images/821-825-E-13th-Ave-Eugene-OR-Building-Photo-1-HighDefinition.png"
inspect_image(image_path)\end{lstlisting}
\textbf{Environment:}
Inspect image: data/viz\_inspector/validation/images/821-825-E-13th-Ave-Eugene-OR-Building-Photo-1-HighDefinition.png

\textbf{Assistant:}\\
Thought: I will analyze the details in the image to identify the shop's name and provide a definitive answer.\\
Code:
\begin{lstlisting}
# Observing the image, I will determine the name of the shop located directly across the street from the highlighted lot.\end{lstlisting}
\textbf{Environment:} (empty)

\textbf{Assistant:}\\
Thought: The shop across the street from the highlighted lot is visible in the image. I will zoom in and analyze the details to identify its name.\\
Code:
\begin{lstlisting}
# The image already provides a clear view. I will analyze the text and features visible in the image to identify the shop.\end{lstlisting}
\textbf{Environment:} (empty)

\textbf{Assistant:}\\
Thought: The shop across the street from the lot for sale is visible in the image. Its name can be determined by observing the signage on the building directly opposite the lot.\\
Code:
\begin{lstlisting}
# Based on the image, the shop's name can be derived from the signage visible across the street. No 
additional processing is needed.\end{lstlisting}
\textbf{Environment:} (empty)

\textbf{Assistant:}\\
Thought: The name of the shop across the street is visible in the image. It is "Starbucks Coffee".\\
Code:
\begin{lstlisting}
submit_final_answer("Starbucks Coffee")\end{lstlisting}
\textbf{Environment:} Final answer: \textbf{Starbucks Coffee}
\\

         \hline
    \end{tabular}
    \caption{Output example 3 (cont.)}
    \label{tab:output-example-3-cont}
\end{table*}

\begin{table*}
    \it
    \small
    \begin{tabular}{|p{\linewidth}|}
    \hline
         \multicolumn{1}{|c|}{\bf DynaSaur (w/ Incentive):} \\
\textbf{Assistant:}\\
Thought: I will use OCR to identify the name of the shop across the street from the lot marked as 'For Sale' in the image.\\

\textbf{... (Omitted for brevity. The agent repeatedly used \texttt{pytesseract} to extract the street name but failed to retrieve any text.) ...}

\textbf{Assistant:}\\
Thought: I will attempt a different approach by using an external search to identify the businesses in the area based on the address and image context.\\
Code:
\begin{lstlisting}
# I will search for businesses in the area of '821-825 E 13th Ave, Eugene, OR' to identify the shop across the street.\\
query = "businesses near 821-825 E 13th Ave, Eugene, OR"\\
informational_web_search(query)\end{lstlisting}
\textbf{Environment:}
A Google search for 'businesses near 821-825 E 13th Ave, Eugene, OR' found 9 results: \textbf{... (Omitted for brevity) ...}
\\

\textbf{Assistant:}\\
Thought: I will refine the search to focus on identifying the specific shop across the street from the lot for sale.\\
Code:
\begin{lstlisting}
# Searching for the specific shop across the street from the lot for sale in Eugene, OR.\\
query = "shop across the street from 821-825 E 13th Ave, Eugene, OR"\\
informational_web_search(query)
\end{lstlisting}
\textbf{Environment:}
A Google search for 'shop across the street from 821-825 E 13th Ave, Eugene, OR' found 10 results: \textbf{... (Omitted for brevity) ...}
\\

\textbf{Assistant:}\\
Thought: Based on the search results, it seems Espresso Roma is a notable shop near the specified location. I will confirm this by cross-referencing additional sources.\\
Code:
\begin{lstlisting}
navigational_web_search("Espresso Roma 825 E 13th Ave Eugene OR")\end{lstlisting}
\textbf{Environment:}
"Address: https://www.yelp.com/biz/espresso-roma-eugene
Title: ESPRESSO ROMA - Updated February 2025 - 64 Photos \& 138 Reviews - 825 E 13th Ave, Eugene, Oregon - Coffee \& Tea - Phone Number - Yelp
Viewport position: Showing page 1 of 2. \textbf{... (Omitted for brevity) ...}"
\\

\textbf{Assistant:}\\
Thought: Based on the evidence gathered, the shop across the street is likely Espresso Roma. I will finalize the answer.\\
Code:
\begin{lstlisting}
submit_final_answer("Espresso Roma")\end{lstlisting}
\textbf{Environment:} Final answer: \textbf{Espresso Roma}
\\
         \hline
    \end{tabular}
    \caption{Output example 3 (cont.)}
    \label{tab:output-example-3-cont-cont}
\end{table*}

\end{document}